\documentclass{article}

\usepackage{microtype}
\usepackage{graphicx}
\usepackage{subfigure}
\usepackage{booktabs} 

\usepackage{hyperref}

\usepackage[accepted]{icml2025}

\usepackage{amsmath}
\usepackage{amssymb}
\usepackage{mathtools}
\usepackage{amsthm}
\usepackage{bbm}

\usepackage{multirow}
\usepackage{booktabs}
\usepackage{graphicx}   
\usepackage{subcaption} 
\usepackage{xspace}
\usepackage[capitalize,noabbrev]{cleveref}

\usepackage{wrapfig}

\theoremstyle{plain}

\theoremstyle{definition}

\theoremstyle{remark}

\usepackage[textsize=tiny]{todonotes}
\usepackage{bm}

\icmltitlerunning{Submission and Formatting Instructions for ICML 2025}

\definecolor{fire}{rgb}{1.0, 0.5, 0.0}

\newcommand{\proj}{LLM-BP\xspace}
\newcommand{\GG}{\mathcal{G}}
\newcommand{\VV}{\mathcal{V}}
\newcommand{\EE}{\mathcal{E}}

\newcommand{\XX}{\boldsymbol{X}}

\newcommand{\hh}{\boldsymbol{h}}

\begin{document}

\twocolumn[
\icmltitle{Generalization Principles for Inference over Text-Attributed Graphs  \\
with Large Language Models}




\begin{icmlauthorlist}
\icmlauthor{Haoyu Wang}{gt}
\icmlauthor{Shikun Liu}{gt}
\icmlauthor{Rongzhe Wei}{gt}
\icmlauthor{Pan Li}{gt}
\end{icmlauthorlist}

\icmlaffiliation{gt}{Department of Electrical and Computer Engineering, Georgia Institute of Technology, Atlanta, GA, USA}

\icmlcorrespondingauthor{Haoyu Wang}{haoyu.wang@gatech.edu}
\icmlcorrespondingauthor{Pan Li}{panli@gatech.edu}

\icmlkeywords{Machine Learning, ICML}

\vskip 0.3in
]



\printAffiliationsAndNotice{} 

\begin{abstract}
Large language models (LLMs) have recently been introduced to graph learning, aiming to extend their zero-shot generalization success to tasks where labeled graph data is scarce. Among these applications, inference over text-attributed graphs (TAGs) presents unique challenges: existing methods struggle with LLMs' limited context length for processing large node neighborhoods and the misalignment between node embeddings and the LLM token space. To address these issues, we establish two key principles for ensuring generalization and derive the framework \proj accordingly: (1) \textbf{Unifying the attribute space with task-adaptive embeddings}, where we leverage LLM-based encoders and task-aware prompting to enhance generalization of the text attribute embeddings; (2) \textbf{Developing a generalizable graph information aggregation mechanism}, for which we adopt belief propagation with LLM-estimated parameters that adapt across graphs. Evaluations on 11 real-world TAG benchmarks demonstrate that \proj significantly outperforms existing approaches, achieving 8.10\% improvement with task-conditional embeddings and an additional 1.71\% gain from adaptive aggregation. The code\footnotemark[2] and task-adaptive embeddings\footnotemark[3] are publicly available. 

\footnotetext[2]{\scriptsize \url{https://github.com/Graph-COM/LLM_BP}}
\footnotetext[3]{\scriptsize \url{https://huggingface.co/datasets/Graph-COM/Text-Attributed-Graphs}}.
\end{abstract}

\section{Introduction} \label{sec:intro}

Inspired by the remarkable generalization capabilities of foundation models for text and image data~\cite{achiam2023gpt, liu2021swin, radford2021learning}, researchers have recently explored extending these successes to graph data~\cite{liu2023towards, mao2024graph, zhao2023gimlet, fan2024graph, he2023harnessing}, aiming to develop models that generalize to new or unseen graphs and thereby reduce reliance on costly human annotation~\cite{li2024glbench, chen2024text, feng2024taglas, li2024teg}. Among various types of graph data, \emph{text-attributed graphs} (TAGs) have found a wide range of applications. These graphs combine both topological relationships and textual attributes associated with each node, which naturally arises in recommendation systems (where user and item nodes may have textual descriptions or reviews)~\cite{bobadilla2013recommender}, academic graphs (where publications include extensive textual metadata)~\cite{mccallum2000automating, giles1998citeseer}, and financial networks (where transactions and accounts come with textual records)~\cite{kumar2016edge, kumar2018rev2}. Given the labeling challenges posed by cold-start problems in recommendation systems or fraud detection in financial networks, methods that can operate with limited labeled data are crucial. In particular, robust zero-shot node labeling across unseen TAGs has become an area of great interest.

\begin{figure*}[t]
    \centering
    \includegraphics[width=0.99\textwidth]{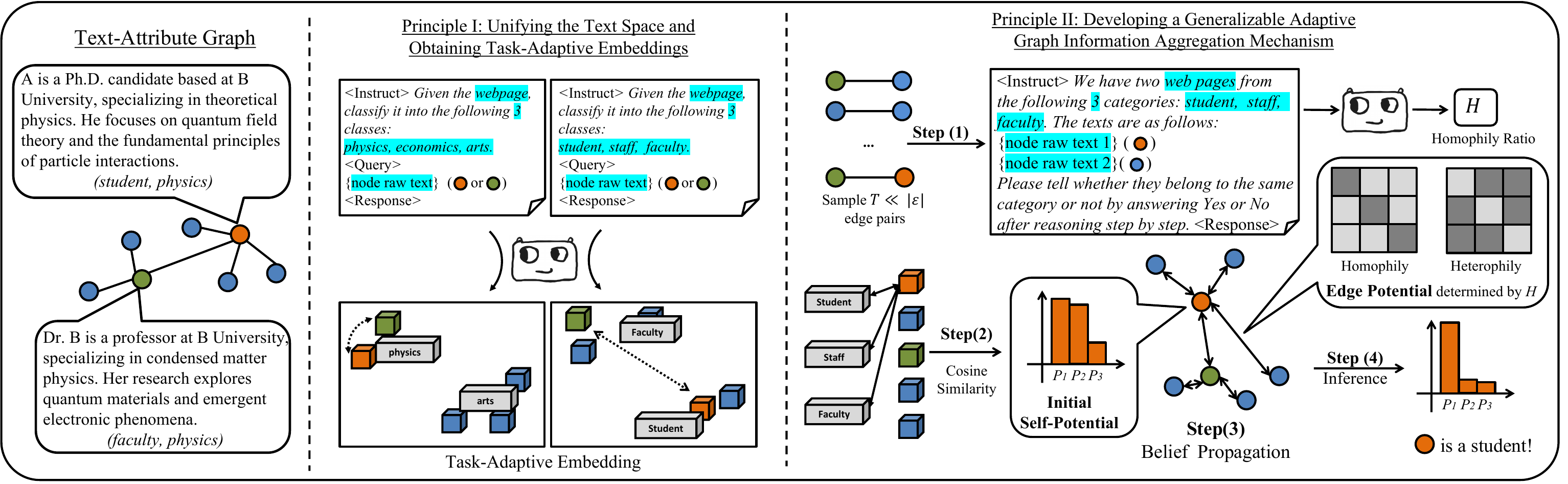}
    \vspace{-0.2cm}
    \caption{The two generalization principles and the framework of LLM-BP.}
    \vspace{-0.4cm}
    \label{fig:pipe}
\end{figure*}

Numerous studies have been dedicated to inference tasks on TAGs. Early efforts have primarily focused on adapting pre-trained language model (LM) encoders~\cite{li2024zerog, fang2024uniglm}, sometimes in combination with graph neural networks (GNNs)~\cite{hou2022graphmae, velivckovic2018deep}, to incorporate structural information. However, these approaches often struggle to achieve strong generalization performance, largely due to the limited capacity of the underlying models. With the advent of large language models (LLMs)~\cite{kaplan2020scaling,huang2022towards}, researchers have proposed two main strategies for integrating LLMs into TAG inference: 1) \textbf{Direct Node-Text Input.} Here, raw node texts are directly fed into LLMs. This method demonstrates reasonably good zero-shot performance on TAGs when text attributes are highly informative for node labels~\cite{chen2024exploring, li2024similarity}. However, when the textual attributes are insufficient, it becomes necessary to aggregate information from a larger neighborhood in the graph, while this is constrained by the limited context length LLMs can digest and reason over. 2) \textbf{Embedding-Based Integration.} In this approach, node texts and their neighboring structural information are first encoded into compressed embeddings, which are then processed by LLMs~\cite{chen2024llaga,tang2024graphgpt,luo2024enhance,wang2024llms,zhang2024graphtranslator}. Because LLMs are not inherently trained on arbitrary embedding spaces, aligning these embeddings with the LLM's token space is essential - an idea partly inspired by how vision-language models align multimodal data~\cite{radford2021learning, zhai2023sigmoid}. However, unlike the vision-language domain, where large-scale text–image pairs~\cite{schuhmann2022laion} are abundant, the graph domain typically lacks comparable datasets. This scarcity reduces the model’s generalization in practice.

In contrast to prior heuristic approaches, this work aims to design a method from first principles for robust zero-shot generalization on TAGs. Because TAGs are inherently multimodal, the proposed method must simultaneously address potential distribution shifts in both textual attributes and graph structure. Specifically, text attributes can vary widely, for example from scientific papers~\cite{mccallum2000automating} to e-commerce product reviews~\cite{ni2019justifying}. Edge connection patterns can range from homophilic graphs such as citation networks, where papers on similar themes are linked~\cite{giles1998citeseer}, to heterophilic graphs such as webpages, which connect nodes with distinct topics~\cite{mernyei2020wiki}. Moreover, the labeling task itself can shift which requires a task-adaptive approach to process both textual features and network structure. Consequently, the core insight behind our model design is grounded in the following two key principles.

\textbf{Principle I: Unifying the text space and obtaining task-adaptive embeddings.} LLMs offer powerful text-understanding capabilities that naturally unify the textual feature space, and have recently been shown to benefit from task-adaptive prompting~\cite{kong2024gofa}. However, to handle the large-scale graph aggregation discussed later, we require these capabilities to extend beyond raw text to an embedding space. Hence, we propose to adopt LLM-based encoder models such as LLM2Vec~\cite{behnamghader2024llm2vec, li2024making} for text embedding. Although this approach might appear to be a naive extension of smaller LM-based embedding methods (e.g., those relying on SBERT~\cite{reimers2019sentence} or RoBERTa~\cite{liu2019roberta}), we argue that leveraging the decoder-induced encoder structure of LLMs is essential for achieving task-adaptive embeddings. In particular, we introduce a novel prompting strategy that encodes text attributes conditioned on inference-task descriptions, enabling significantly improved zero-shot inference - an ability not readily achieved by smaller LM-based embeddings.

\textbf{Principle II: Developing a generalizable adaptive graph information aggregation mechanism.} Graph structure determines the node neighboring relationships and thus the information aggregation from which nodes may benefit the inference. Inspired by the belief propagation (BP) algorithm~\cite{murphy2013loopy} that gives the optimal statistical inference over pairwise correlated random variables, we propose to regard the graph as a Markov Random Field (MRF), each node as a random variable, and mimic BP to aggregate information for node label inference. Because BP is rooted in basic mathematical principles, this approach is widely generalizable. Algorithmic adaptivity across different TAGs hinges on estimating the coupling coefficients in the graphs, which can be done by having LLMs analyze the attributes of sampled pairs of connected nodes. Moreover, this BP-inspired approach naturally adapts to varying levels of text attribute quality: nodes with higher-quality text attributes present greater influence on their neighbors, and vice versa.

By applying the two principles outlined above, we propose our new strategy, \proj, for zero-shot inference over TAGs. \proj does not require any training or fine-tuning. We evaluate \proj on 11 real-world TAG datasets from various domains, including citation networks~\cite{mccallum2000automating, giles1998citeseer, sen2008collective}, e-commerce~\cite{ni2019justifying}, knowledge graphs~\cite{mernyei2020wiki}, and webpage networks~\cite{craven1998learning}, covering both homophilic and heterophilic graph structures. 

Experimental results demonstrate the effectiveness of \proj. Notably, our task-conditional embeddings (Principle I) improve performance by $8.10\%$ on average compared to the best LM-based encoders. In addition, our BP-inspired aggregation mechanism (Principle II) provides an extra $1.71\%$ performance gain with our embeddings, demonstrating strong generalization across both homophilic and heterophilic TAGs.  Our experiments also reveal that current methods aligning graph-aggregated embeddings to LLM token spaces significantly underperform approaches that simply use smaller LM encoders without even incorporating graph structures. This outcome indicates that the primary source of generalization in these methods is the smaller LM’s text embeddings, rather than LLM-based reasoning on embeddings. It reinforces our earlier argument that limited training data hinders effective alignment in this context, urging caution for future work considering this strategy.

\section{Related Works}
Here, we briefly review existing methods by examining how they enable model generalization across TAGs. 

\textbf{Tuning Smaller LM Encoders.} 
These methods typically rely on a source-domain graph for training. Notable works include ZeroG~\cite{li2024zerog} that tunes SBert~\cite{reimers2019sentence} on source datasets to align class-description embeddings with node text, thereby enhancing zero-shot performance on target datasets. Another approach, UniGLM~\cite{fang2024uniglm}, fine-tunes BERT~\cite{kenton2019bert} using contrastive learning on source datasets to yield a more generalizable encoder. GNNs trained with UniGLM embeddings in a supervised manner outperform models that directly adopt LM embeddings.


\textbf{Training GNNs for Generalization.} These methods focus on leveraging graph structure in a generalizable manner. Among them, graph self-supervised learning~\cite{liu2022graph} is particularly common for producing representations without labeled data, often employing contrastive learning or masked modeling~\cite{velivckovic2018deep, hou2022graphmae, zhao2024all}. GraphMOE is a more recent technique inspired by the success of mixture-of-experts~\cite{shazeer2017outrageously}, pre-training parallel graph experts targeting different structures or domains~\cite{hou2024graphalign, liu2024one, xia2024anygraph, qin2023disentangled}. 
Others also consider LM-GNN co-training including \cite{he2024unigraph,zhu2024graphclip} that also follow a constrastive learning idea. 
Note that, however, all these methods still require training.  

In contrast to the above effort that adopts smaller LM encoders, works that involve the use of LLMs are reviewed in the following and may achieve better generalization. More related works including LLM-based data augmentations for GNN training for generalization and LLMs for other graph reasoning tasks can be found in Appendix.~\ref{sec:app_more_related_works}.



\textbf{LLMs with Node-Text Input.} LLMs being directly fed with raw node texts demonstrates strong zero-shot ability on TAGs~\cite{chen2024exploring, huang2024can, li2024similarity}. However, they suffer from the limitation of not being able to incorporate graph structural information. 

\textbf{LLMs with Graph-Embedding Input.} With smaller LM-encoded node embeddings, various strategies integrate graph structure by aggregating these embeddings, such as neighborhood-tree traversal or concatenating the averaged embeddings from different hops~\cite{chen2024llaga, tang2024graphgpt, luo2024enhance}, or via pre-trained GNNs~\cite{zhang2024graphtranslator, wang2024llms}. As mentioned earlier, these methods rely on aligning embeddings with the LLMs' token space. For instance, LLaGA~\cite{chen2024llaga} trains a simple MLP on citation networks and~\cite{wang2024llms} employs a linear projector on the ogbn-Arxiv~\cite{hu2020open} dataset, both using the next-token prediction loss, while \cite{tang2024graphgpt} adopts self-supervised structure-aware graph matching as the training objective. However, due to limited TAG-domain data, the space alignment in these methods often remains undertrained, leading to degraded performance.

\textbf{Multi-Task Graph Foundation Models.} More ambitious studies aim to generalize across various graph-related tasks within a single framework. Notable approaches include graph prompting~\cite{liu2023graphprompt}, which introduces ``prompting nodes'' to transform diverse graph tasks into a unified format. These frameworks then train GNNs to address the tasks~\cite{li2024zerog, liu2023one, liu2024one} or further integrate LLMs~\cite{yang2024gl, kong2024gofa}. Although these works are impressive, they still fail to achieve zero-shot performance comparable to those methods that focus on specific graph data domains. 



\section{Generalization Principles for \proj}

\subsection{Notations and Problem Formulation} 

Let $(\GG, X, Y)$ represent a TAG of interest, where $\GG(\VV, \EE)$ denotes the graph structure, $\VV$ is the node set of size $n$, and $\EE$ is the edge set. The node textual attributes are represented as $X = \{X_1, ..., X_n\}$, and each node belongs to one of $c$ classes, with labels given by $Y=\{y_1,y_2,...,y_n\} \in [c]^n$. 

The objective is to infer the labels of nodes in TAGs based on the node attributes and graph structure. This study primarily focuses on the \textbf{zero-shot} setting, where no labeled data are assumed to be available in advance. Additionally, a \textbf{few-shot} setting is considered, where $k$ labeled nodes are known for each class. Due to space limitations, results for the few-shot setting are provided in Appendix~\ref{sec:app_few_shot}.  

\subsection{Motivation and the Overall Framework}
 LLMs are commonly used as decoders for next-token prediction. While LLMs excel at processing natural language inputs, they are not inherently compatible with graph data. Recently, some studies have explored methods to integrate graph data into LLMs, primarily for reasoning tasks~\cite{perozzi2024let, zhang2024can, tang2024grapharena}. 

In the context of TAGs, accurate node label inference relies on effectively combining the attributes of multiple nodes, especially when a node's individual attributes are insufficient to determine its label. However, as noted earlier, LLMs are constrained by limited context windows, making it challenging to process all attributes from the potentially large set of connected nodes. Traditional approaches to compressing graph structural information involve creating embeddings, such as using GNNs to aggregate information from the target node's neighbors. While effective, these embedding methods do not seamlessly integrate with LLM inputs and often require non-trivial training effort to align the LLMs' token space with the node embedding space~\cite{chen2024llaga, wang2024llms, tang2024graphgpt}. 

Our approach, \proj, does not confine LLMs to their traditional usage. We first leverage their capabilities to generate task-adaptive node embeddings. Then, instead of requiring LLMs to directly process these embeddings, LLMs are further employed to analyze graph data and provide generalizable guidance in aggregating these embeddings. These two steps are to match the two generalization principles proposed in Sec.~\ref{sec:intro}. Classification is ultimately performed by computing the cosine similarity between the final node embeddings $\mathbf{h}^X=[h_1^X,...,h_n^X]$ and candidate class embeddings $\mathbf{q}^C=[q_1^C,...,q_c^C]$. In the zero-shot setting, class embeddings are generated as follows: we randomly sample $l \ll n$ nodes and employ LLMs to infer their labels. The embeddings of sampled nodes form distinct clusters based on LLMs' prediction. We compute the average embedding of the embeddings closest to the cluster center to obtain the class embedding. 
In the few-shot setting, class embeddings are obtained by averaging the embeddings of labeled nodes within each class. See Appendix.~\ref{sec:app_llm_bp_implementation_details} for details.

\subsection{Principle I: Task-Adaptive Node Embeddings $\mathbf{h}^X$}

Creating generalizable text embeddings is no longer a significant challenge. Even smaller LM encoders, such as SBert~\cite{reimers2019sentence}, are capable of achieving this. Indeed, most existing works utilize these encoders to generate initial node embeddings for TAGs~\cite{chen2024text,chen2024llaga, tang2024graphgpt, wang2024llms}. However, for these embeddings to be directly usable for label prediction without the need for additional transformation models, it is crucial to incorporate task-specific information. In other words, the embeddings must be tailored to the specific task, resulting in what we term task-adaptive embeddings.

Achieving task adaptivity, however, presents a notable challenge. Smaller LM encoders lack the expressive power necessary to encode nuanced task-specific information. This limitation motivates our adoption of LLM-induced encoders, driven by the emergent capabilities of LLMs in contextual learning~\cite{sahoo2024systematic, chen2023unleashing}.

There have been recent advancements in extending LLMs to generate text embeddings~\cite{behnamghader2024llm2vec, muennighoff2022mteb}. In our approach, we utilize a form of LLM2Vec~\cite{behnamghader2024llm2vec}, which transforms LLM decoders into encoders via retaining the unidirectional attention mechanism. Following the methodology 
 in~\cite{li2024making}, we extract the output embedding of $\langle\text{response}\rangle$ - the token positioned immediately after the inputs - as the text embedding for the input node attributes.

To embed task-specific information into node embeddings, we propose a prompting strategy structured with the following template:
\begin{align}
\label{eq:task_adaptive_llm2vec}
\resizebox{0.43 \textwidth}{!}{$\langle\text{Instruct}\rangle \{\text{task\_description}\} \{\text{class\_info}\} \langle\text{query}\rangle {X_i} \langle\text{response}\rangle$}.
\end{align}

Here, $\langle \cdot \rangle$ encloses specific tokens. The task details are described in $\{\text{task\_description}\}$, and $\{\text{class\_info}\}$ contains the basic information of each class. An example is given in Fig.~\ref{fig:pipe}.
The class information serves as a crucial contextual enhancement, enabling LLMs to generate embeddings in a conditioned manner. 
For more detailed on the class-conditional prompt for each dataset used in this study, refer to Appendix.~\ref{sec:app_llm_bp_implementation_details} and ~\ref{sec:app_prompt_llm2vec_task_description}.

\subsection{Principle II: Generalizable Graph  Aggregation} \label{sec:principle2}
Graph structures can provide essential correlated information for node label inference by characterizing the relationships between node attributes and labels~\cite{zhu2003semi,kipf2016semi,velivckovic2017graph,hamilton2017inductive,zhu2020beyond,wei2022understanding}.

 Specifically, we may consider each node's label and attributes as random variables, and each edge as a coupling between them for connected node pairs. The fundamental BP algorithm enables principled statistical inference over this set of correlated random variables~\cite{murphy2013loopy}. Since BP is inherently agnostic to the application domain of the TAG, emulating BP offers a mechanism to aggregate correlation information encoded in the graph structure across domains.

\textbf{Markov Random Field Modeling} We consider the joint probability distribution $\mathbb{P}_\mathcal{G}(Y, X)$ over the graph where $Y$ and $X$ denotes the random variables of node labels and attributes, respectively. 
In $\mathbb{P}_\mathcal{G}(Y, X)$, the distribution over the node labels given the graph structure is denoted as
\begin{align}
\mathbb{P}_\mathcal{G}(Y)
= \frac{1}{Z_{\mathbf{Y}}} \prod_{i \in \mathcal{V}} \phi_i(y_i) \prod_{(i,j) \in \mathcal{E}} \psi_{ij}(y_i, y_j).
\end{align}
Here $\phi_i(y_i)$ denotes the unary potential for node $i$, $\psi_{ij}(y_i, y_j)$ is the edge potential capturing the correlation between labels $y_i$ and $y_j$ of adjacent nodes, and $Z_{Y}$ is the normalization constant.  For node attributes, MRF modeling assumes that each node’s attributes are conditionally independent of others given the node labels, which can be characterized by the distribution:  
\begin{align}
\mathbb{P}_\mathcal{G}(X \mid Y) 
= \prod_{i \in \mathcal{V}} \mathbb{P}_\mathcal{G}(X_i \mid y_i) 
= \prod_{i \in \mathcal{V}} \varphi_{y_i}(X_i)
\end{align}
where $\varphi_{y_i}(X_i)$ captures the likelihood of having node $i$’s attributes $X_i$ given its label $y_i$. 

The proposed modeling approach is highly adaptive, as it can capture the varying graph connectivity patterns across different TAGs through interpretable edge potentials. For instance, $\psi_{ij}(y_i, y_j)$ represents the unnormalized likelihood that nodes with labels $y_i$ and $y_j$ are connected. This formulation naturally incorporates the modeling of graph homophily and heterophily: $\psi_{ij}(y_i, y_i) > \psi_{ij}(y_i, y_j)$ indicates homophily, while $\psi_{ij}(y_i, y_i) < \psi_{ij}(y_i, y_j)$ reflects heterophily. Furthermore, $\varphi_{y_i}(X_i)$ enables the model to account for variations in the quality of text attributes (w.r.t. their indicative power for the labels) across different TAGs, further enhancing its adaptivity. For node classification, we can infer $\mathbb{P}_\mathcal{G}(Y \mid X) \propto \prod_{i \in \mathcal{V}} \varphi_{X_i}(y_i) 
\prod_{(i,j) \in \mathcal{E}} \psi_{ij}(y_i, y_j)$ where $\varphi_{X_i}(y_i) = \varphi_{y_i}(X_i)\phi_i(y_i)$.

\begin{algorithm}[t]
\caption{LLM-BP}
\label{alg:llm_bp}
\begin{algorithmic}[1]
\INPUT TAG $(\GG, \XX)$
\OUTPUT Class label prediction $\{\hat{y}_i\}_{i\in[n]}$
\STATE $\hh^{X}$ $\leftarrow$ Task-adaptive encoding following Eq.~\eqref{eq:task_adaptive_llm2vec}
\IF{zero-shot}
\STATE Sample $l \ll n$ nodes, infer labels with LLMs,
\STATE Nodes clusters based on LLM prediction,
\STATE $\mathbf{q}^{C}$ $\gets$ Average embedding of samples near center,
\ELSIF{few-shot}
\STATE $\mathbf{q}^{C}$ $\gets$ Average embedding of $k$ samples per class,
\ENDIF
\STATE Estimate $\psi_{ij}(y_i, y_j)$ by employing the LLM to analyze the graph data (e.g., using Eq.~\eqref{eq:bp} based on the estimated homophily level $r$.)
\STATE Initialize $p^{(0)}(y_i)$ $\gets$ Eq.~\eqref{eq:node_potential} and $m^{(0)}_{i \to j} (y_j) = 1$
\STATE Run LLM-BP (Eq.~\eqref{eq:message_passing}) for $L$ iterations or its approximation (Eq.~\eqref{eq:bp_appr}) for single iteration
\STATE $\hat{y_i}$ $\gets$ $\arg \max_{y_i} \log p_i^{(k)}(y_i;x_i)$

\end{algorithmic}
\end{algorithm}

\textbf{Belief Propagation} 
 Exact inference for $\mathbb{P}_{\GG}(Y|X)$ is intractable in large-scale graphs with cycles~\cite{koller2009probabilistic}. In practice, loopy belief propagation (LBP) is often used to conduct an approximate inference~\cite{murphy2013loopy}, which follows: Initialize the distributions $p_j^{(0)}(y_j)\propto\varphi_{X_i}(y_i)$ and $m_{i \to j}^{(0)}(y_j) = 1/c$ for all $i,j\in\VV$. For $k=1,2,...,L$, we do 
 \begin{align}
 \label{eq:message_passing}
 \log m_{j \to i}^{(k)}(y_i) \cong &\, \text{LSE}_{y_j}[\log \psi_{ij}(y_i,y_j) + \\ &
\log p_j^{(k-1)}(y_j) - \log m_{i \to j}^{(k-1)}(y_j)], \nonumber \\
 \log p_i^{(k)}(y_i) \cong &\log p_i^{(0)}(y_i) + \sum_{j \in \mathcal{N}(i)} \log m_{j \to i}^{(k)}(y_i),  \nonumber
\end{align} where $\cong$ denotes the equality with difference up-to a constant. LSE stands for the log-sum-exp function: 
$\text{LSE}_{y_j} [f(y_i, y_j)] = \log \left[ \sum_{y_j} \exp (f(y_i, y_j)) \right]$. The final $\arg\max_{y_i} p_i^{(k)}(y_i)$ gives the label prediction. 
Detailed derivation can be found in Appendix.~\ref{sec:app_detailed_derivation}. 

\textbf{\proj} To execute the above LBP algorithm, we need to specify several components based on the TAG. First, $p_i^{(0)}(y_i)$ represents the distribution of the label $y_i$ given the observed attributes $X_i$ alone, which can be estimated using normalized cosine similarities: 
\begin{align}
\label{eq:node_potential}
p_i^{(0)}(y_i) = \text{softmax}(\{\cos(h^{X}_i, h^{C}_k)/\tau\}_{k\in[c]})
\end{align}

where $h_i^{X}$ and $h_k^{C}$ denote node $i$'s class-conditional embedding and class $k$'s embedding given by the LLM encoder as discussed in the previous section. 
$\cos(\cdot)$ denotes cosine similarity and $\tau$ is the temperature hyper-parameter. 

Second, we characterize the edge potentials $\psi_{ij}(y_i, y_j)$. We employ an LLM agent to assess the homophily level of the TAG. Specifically, we uniformly at random sample $T$ connected node pairs ($T\ll |\EE|$), and for each pair, we prompt the LLM to determine whether the two nodes belong to the same class based on their attributes, as illustrated in Fig.~\ref{fig:pipe}. The ratio of ``Yes'' responses, denoted by $r$ is used to set 
\begin{align}
\label{eq:bp}
\ \psi_{ij}(y_i, y_j) =
\begin{cases}
r, & \text{if } y_i = y_j \\
1-r, & \text{if } y_i \neq y_j
\end{cases},
\end{align}
Note that a more complex $\psi_{ij}(y_i, y_j)$ can be adopted by estimating the edge probabilities between any two classes. However, we choose the homophily level as a proof of concept. LLMs can provide a reasonably accurate estimation of the homophily level, as pairwise comparisons are typically much simpler tasks compared to full-scale classification.

\cite{wei2022understanding} demonstrated that linear propagation can approximate a single iteration of LBP when feature quality is limited. Based on this insight, we adopt the following approximate LBP formulation (denoted as BP appr.):
\begin{align}
 \log p_i^{(1)}(y_i) \cong &\log p_i^{(0)}(y_i)+
 \label{eq:bp_appr}\\
 &\text{sgn}(\log \frac{r}{1-r})\sum_{j \in \mathcal{N}(i)} \log p_j^{(0)}(y_i),  \nonumber    
\end{align}
where the homophily level $r$ influences the sign of the log-likelihood aggregation from neighboring nodes. We summarize the overall pipeline in Algorithm.~\ref{alg:llm_bp}

\section{Experiments} 
In this section, we evaluate \proj based on its two design principles, with a primary focus on zero-shot node classification tasks. Evaluations of few-shot node classification and link prediction tasks are provided in Appendix.~\ref{sec:app_few_shot}~\ref{sec:app_link_prediction}. First, we demonstrate the effectiveness of task-adaptive encoding and identify issues with existing methods that rely on aligning node embeddings with the LLM token space. Second, we validate the effectiveness of the proposed BP algorithm. Finally, we present the end-to-end performance of \proj, comparing it to state-of-the-art baselines. We first introduce the datasets and baselines used in the study:

\textbf{Datasets} As listed in Table~\ref{tab:datasets}, we selected eleven real-world TAG datasets that encompass a variety of text domain shifts, including citation networks, e-commerce data, knowledge graphs, and webpage networks, which cover both homophily and heterophily structures. For more details, see Appendix~\ref{sec:app_datasets}. 

\begin{table}
\centering
\resizebox{0.46 \textwidth}{!}{\begin{tabular}{@{}cccc@{}}
\toprule
 Dataset & Text Domain & Graph Structure \\ \midrule
Cora~\cite{mccallum2000automating}& CS Publication & Homopholic \\
Citeseer~\cite{giles1998citeseer} & CS Publication & Homopholic \\
Pubmed~\cite{sen2008collective} & Medical Publication & Homopholic \\
History~\cite{ni2019justifying} & History Books & Homopholic \\
Children~\cite{ni2019justifying} & Children Literature & Homopholic \\
Sportsfit~\cite{ni2019justifying} & Sports Goods & Homopholic \\
Wikics~\cite{mernyei2020wiki} & Knowledge Graph & Homopholic \\
Cornell~\cite{craven1998learning} & School Webpage & Heterophilic \\
Texas~\cite{craven1998learning} & School Webpage & Heterophilic \\
Wisconsin~\cite{craven1998learning} & School Webpage & Heterophilic \\
Washington~\cite{craven1998learning} & School Webpage & Heterophilic \\ \bottomrule
\end{tabular}}
\vspace{-0.cm}
\caption{TAG Datasets selected in experiments.}
\vspace{-0.8cm}
\label{tab:datasets}
\end{table}

\textbf{Baselines:} We select representative baselines from all existing categories for model generalization on TAGs:

$\bullet$ \textit{Vanilla LM / LLM Encoders}: including Sentence-BERT (SBert)~\cite{reimers2019sentence}, RoBERTa~\cite{liu2019roberta}, text-embedding-3-large~\cite{openai2024textembedding}, and bge-en-icl~\cite{li2024making}, a state-of-the-art LLM2Vec encoder.

$\bullet$ \textit{Vanilla LLMs}: including GPT-3.5-turbo~\cite{achiam2023gpt} and GPT-4o~\cite{hurst2024gpt}, the latter being among the most advanced LLMs in reasoning. They process raw node texts without incorporating graph structures.

$\bullet$ \textit{Tuning LM Encoder / GNNs}: including ZeroG~\cite{li2024zerog}, UniGLM~\cite{fang2024uniglm} that tune LM encoders, ZeroG is specifically proposed for zero-shot node classification. DGI~\cite{velivckovic2018deep}, GraphMAE~\cite{hou2022graphmae} that perform Graph-SSL are also compared.

$\bullet$ \textit{LLMs with Graph Adapters}: including LLaGA~\cite{chen2024llaga}, TEA-GLM~\cite{wang2024llms}, and GraphGPT~\cite{tang2024graphgpt}, which are the three representative works adopting LLMs with projectors to align compressed node representations with the token space.

$\bullet$ \textit{Multi-Task Graph Foundation Models}: Consisting of OFA~\cite{liu2023one} and GOFA~\cite{kong2024gofa}, which are the state-of-the-art multi-task foundation models.

$\bullet$ \textit{LLMs for Data Augmentation}: referring to LLM-GNN~\cite{chen2023label}, specifically designed for zero-shot node classification, which utilizes LLMs as annotators for pseudo-labels and further train GNNs for inference.

$\bullet$ \textit{Neighborhood Aggregation (NA)}: referring to the training-free method proposed in ~\cite{yang2024gl}, which injects graph structural information into node representations by directly aggregating the averaged neighborhood embeddings.

\textbf{Settings:} Unlike LLM-BP which does not require additional fine-tuning of LLMs, most of the baselines--except from vanilla encoders, LLMs or NA--require fine-tuning. Methods of vanilla encoders and LLM-BP that require sampling nodes to obtain class embeddings under zero-shot settings are repeated 30 times with seed 42 to 71, and the average performance is reported in the following experiment sections. Implementation details for baselines and LLM-BP can be found in Appendix.~\ref{sec:app_implementation_details_baseline}~\ref{sec:app_llm_bp_implementation_details}.

\subsection{Evaluation for Task-Adaptive Node Embedding}

\begin{figure}[t]
    \centering 
    \begin{minipage}{0.46\textwidth}  
        \centering
        \includegraphics[width=\textwidth]{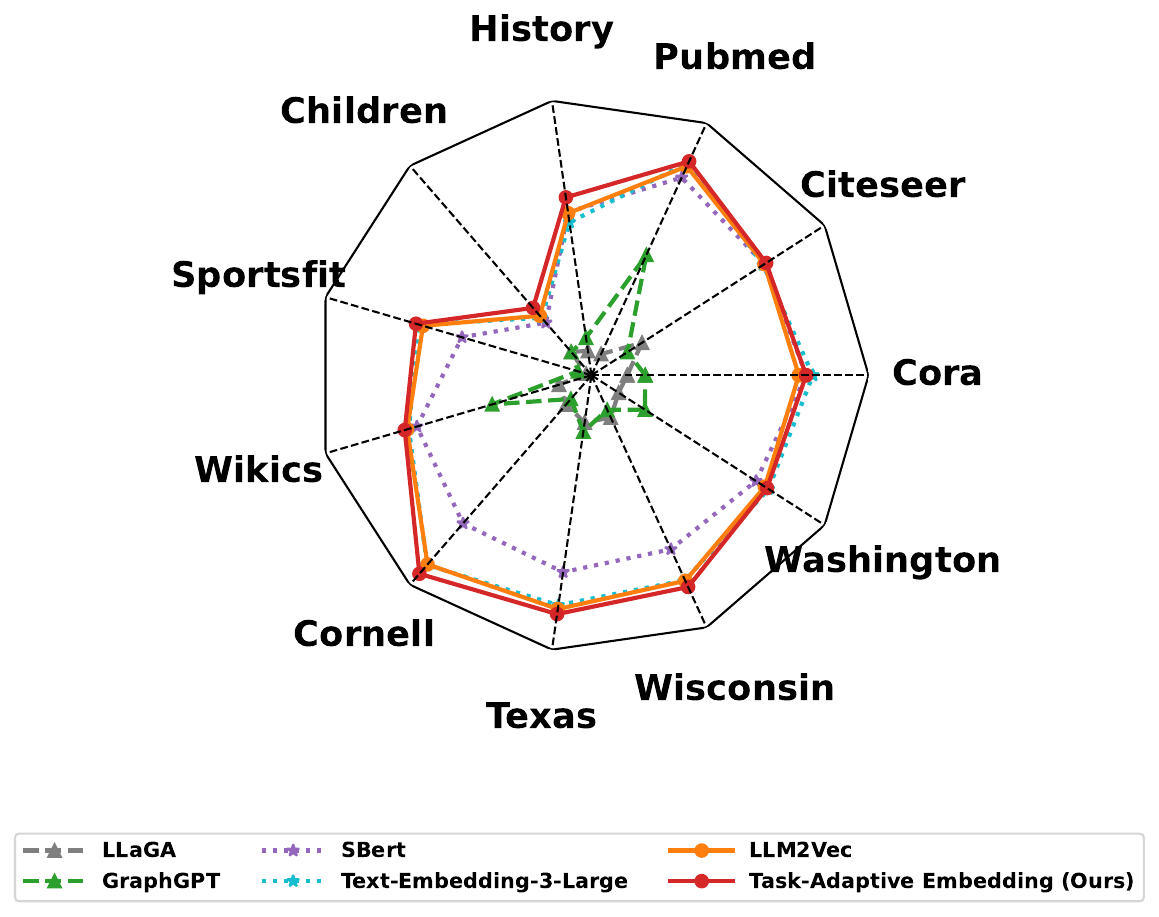}
        
        \caption{Zero-Shot Accuracy of vanilla encoders vs. LLMs-with-Graph-Adapters. All the encoder-based methods do not leverage graph structure information.}
        \label{fig:principle_1}
    \end{minipage}
    \begin{minipage}{0.46\textwidth}  
        \centering
        \begin{minipage}{0.49\textwidth}
            \centering
            \includegraphics[width=\textwidth]{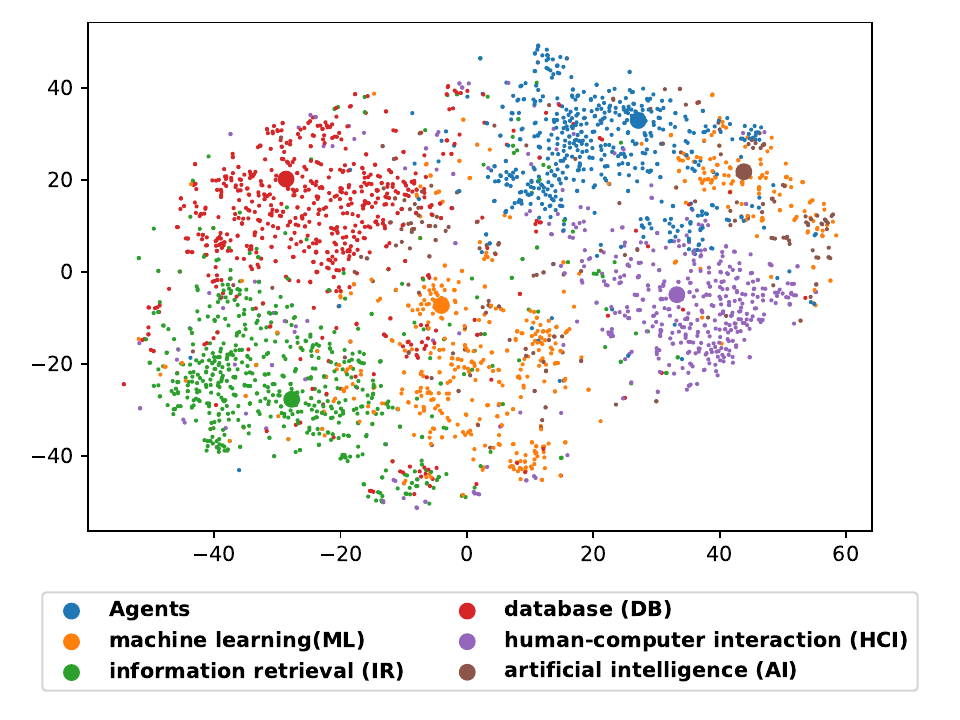}
            \caption*{\scriptsize SBert.}
        \end{minipage}
        \begin{minipage}{0.49\textwidth}
            \centering
            \includegraphics[width=\textwidth]{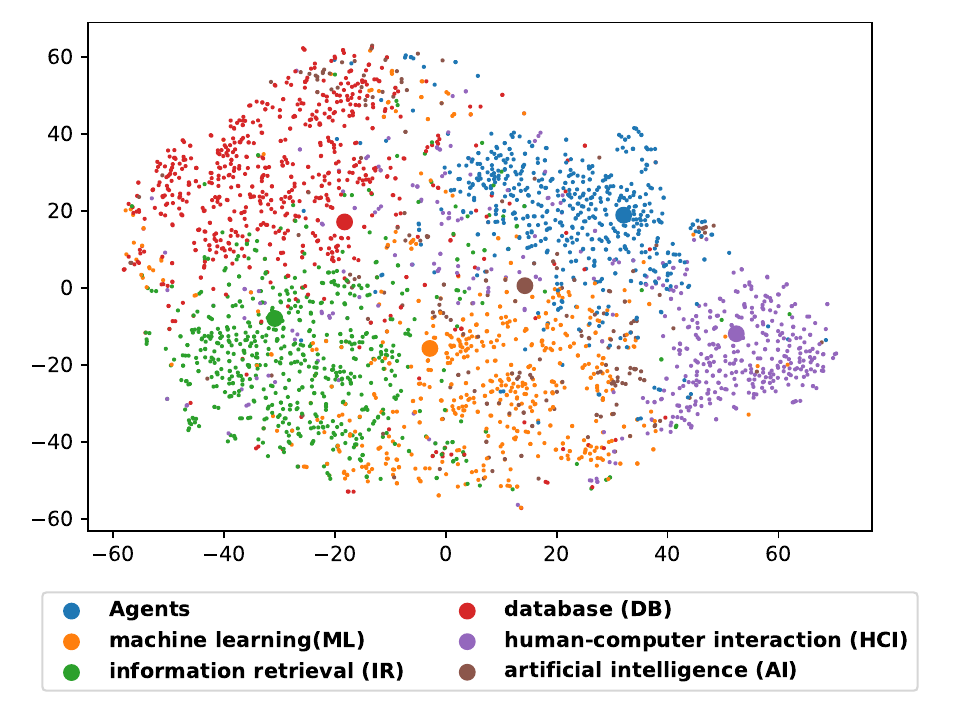}
            \caption*{\scriptsize Text-Embedding-3-Large.}
        \end{minipage}
        
        \begin{minipage}{0.47\textwidth}
            \centering
            \includegraphics[width=\textwidth]{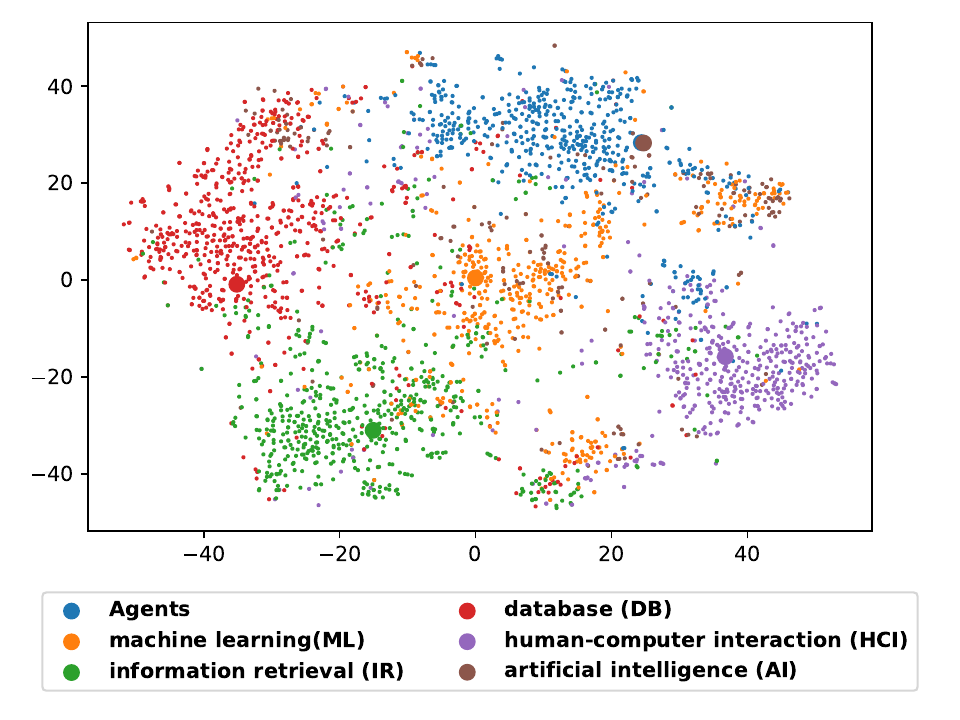}
            \caption*{\scriptsize LLM2Vec.}
        \end{minipage}
        \begin{minipage}{0.47\textwidth}
            \centering
            \includegraphics[width=\textwidth]{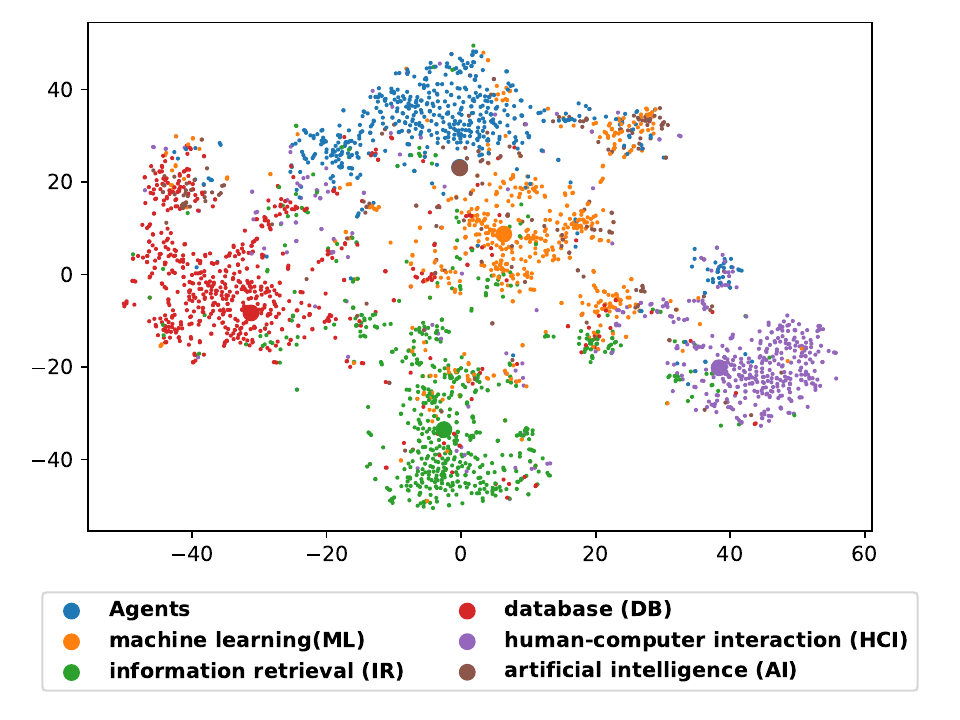}
            \caption*{\scriptsize Task-Adaptive Encoder (Ours).}
        \end{minipage}
        \caption{t-SNE visualization of encoders on Citeseer.}
        \label{fig:tsne}
    \end{minipage}
    \hfill
   
\end{figure}

$\bullet$ \textbf{Exp.1: Ineffectiveness of LLMs w/ Graph Adapters}  

Figure~\ref{fig:principle_1} illustrates the accuracy of encoder-based methods alongside two representative LLMs-with-graph-adapters methods across each dataset. Notably, using text embeddings generated by SBert~\cite{reimers2019sentence} without incorporating graph structural information significantly outperforms both LLaGA~\cite{chen2024llaga} and GraphGPT~\cite{tang2024graphgpt}. These two methods align node representations that combine SBert embeddings with graph information to the LLMs' token space via a projector. This finding suggests that the generalization capabilities of these approaches primarily stem from the pre-trained language model encoders rather than the LLMs' inherent understanding of TAG data. Consequently, future works should exercise caution when adopting this strategy.

$\bullet$ \textbf{Exp.2: Effectiveness of The Task-Adaptive Encoder}

\begin{figure}[t]
    \centering
    \begin{minipage}{0.15\textwidth}
    \captionsetup{labelformat=empty}
        \centering
        \includegraphics[width=\textwidth]{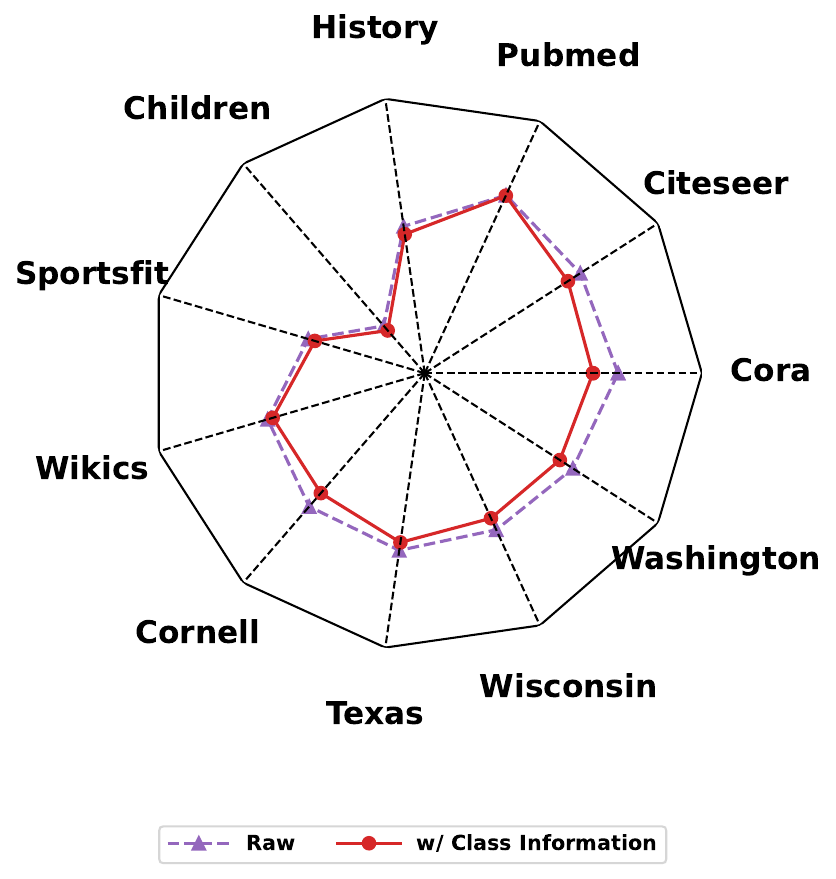}
        \vspace{-0.8cm}
        \caption*{\scriptsize SBert (Encoder).}
    \end{minipage}
    \begin{minipage}{0.15\textwidth}
    \captionsetup{labelformat=empty}
        \centering
        \includegraphics[width=\textwidth]{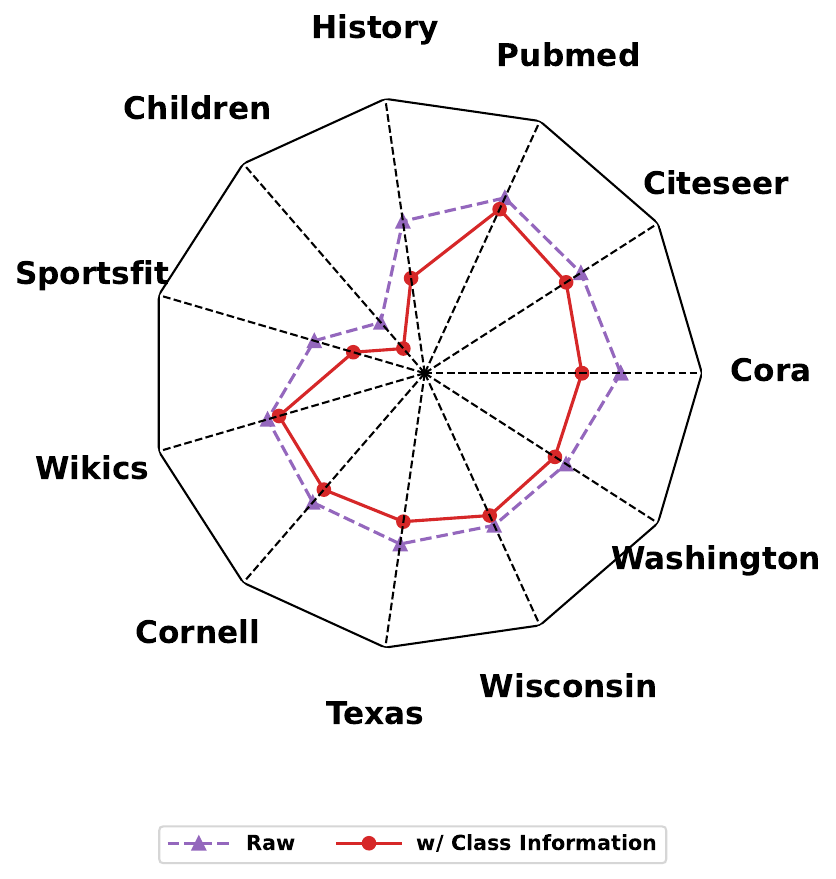}
        \vspace{-0.8cm}
        \caption*{\scriptsize Roberta (Encoder).}
    \end{minipage}
    \begin{minipage}{0.15\textwidth}
    \captionsetup{labelformat=empty}
        \centering
        \includegraphics[width=\textwidth]{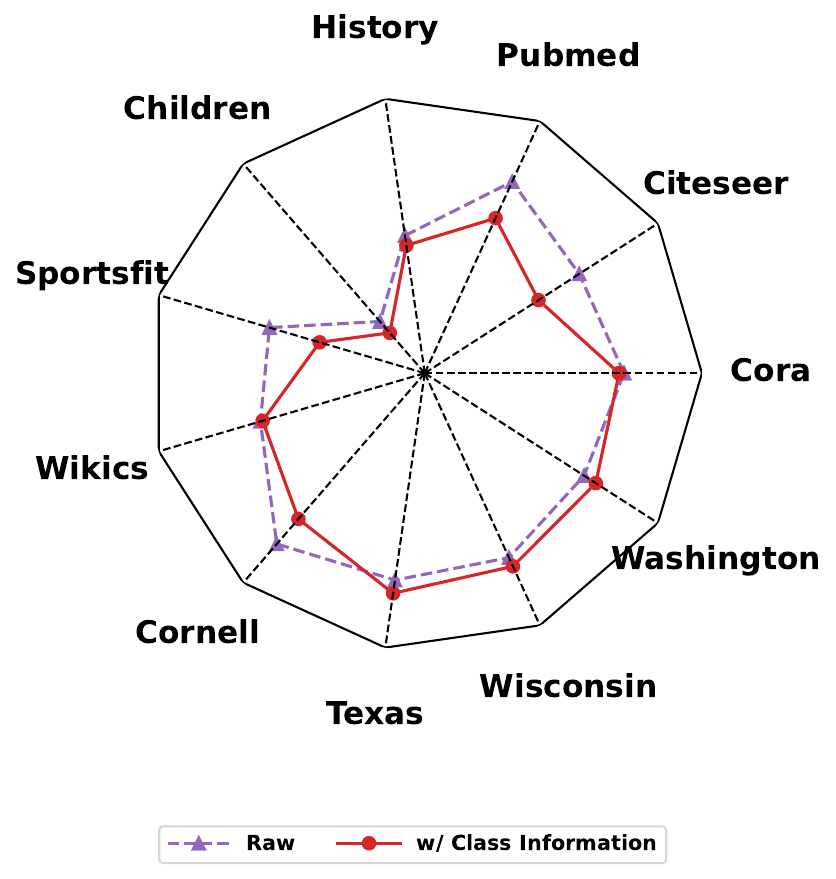}
        \vspace{-0.8cm}
        \caption*{\scriptsize  Text-Embedidng-3-Large.}
    \end{minipage}
    \vspace{-0.2cm}
    \caption{Class information fed into different encoders.}
    \label{fig:class_condition}
\end{figure}
According to Figure~\ref{fig:principle_1}, the task-adaptive encoder achieves the best performance on most of the datasets, enhancing the vanilla LLM2Vec on average by $2.3\%$, highlighting the importance of incorporating task-specific information during encoding. 
To further illustrate this, we use the Citeseer~\cite{giles1998citeseer} dataset as an example and perform t-SNE visualization~\cite{van2008visualizing} on the embeddings derived from the encoders. As shown in Fig.~\ref{fig:tsne}, when provided with class information, the task-adaptive encoder generates embeddings that exhibit tighter clustering for the same class compared to other baselines. The significance test of improvement from task-adatove encoding is provided in Table.~\ref{tab:confidence} in Appendix.~\ref{sec:app_confidence}.

Note that the benefits of class information are observed only in encoders derived from LLM decoders potentially due to their strong contextual learning capabilities. As illustrated in Fig.~\ref{fig:class_condition}, incorporating class information into smaller LM encoders, such as SBert~\cite{reimers2019sentence} or RoBERTa~\cite{liu2019roberta}, may even degrade performance.  
Regarding Text-embedding-3-large~\cite{openai2024textembedding}, the impact of class information remains inconclusive due to the unknown internal mechanisms of the black-box encoder.

\subsection{Generalizable Graph Aggregation}
\begin{figure}[ht]
    \centering
    \begin{minipage}{0.20\textwidth}
    \captionsetup{labelformat=empty}
        \centering
        \resizebox{\textwidth}{!}{\begin{tabular}{@{}cc@{}}
        \toprule
        Graph Type & \# Sampled Edges \\ \midrule
        Citation & 100 \\
        E-Commerce & 100 \\
        Knowledge Graph & 100 \\
        Web Page & 50 \\ \bottomrule
        \end{tabular}}
        \vspace{-0.0cm}
    \end{minipage}
    \begin{minipage}{0.25\textwidth}
    \captionsetup{labelformat=empty}
        \centering
        \includegraphics[width=\textwidth]{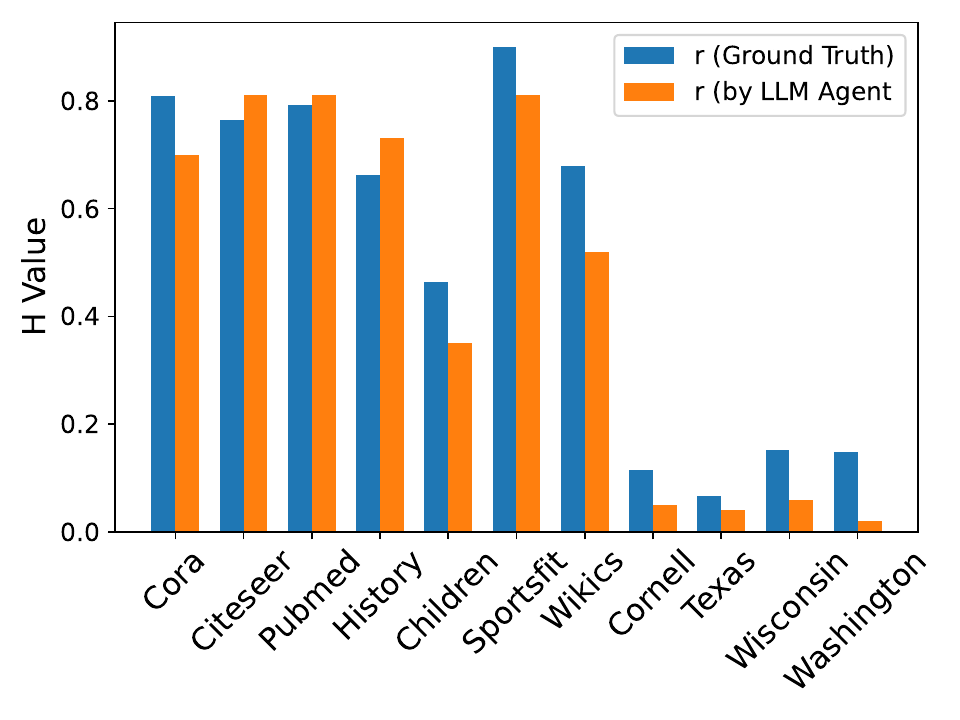}
        \vspace{-0.8cm}
    \end{minipage}
    \vspace{-0.1cm}
    \caption{Left: Number of edges sampled per dataset. Right: GPT-4o-mini's prediction of the homophily level $r$.}
    \label{fig:predict_h}
\end{figure}

\begin{table*}[t]
\setlength{\tabcolsep}{2pt}
\renewcommand{\arraystretch}{1.2}
\resizebox{\textwidth}{!}{
\begin{tabular}{c|cccccccccccccc|cccccccc|cc}
\hline
 & \multicolumn{14}{c|}{Homophilic} & \multicolumn{8}{c|}{Heterophilic} & \multicolumn{2}{c}{Avg Rank} \\ \hline
 & \multicolumn{6}{c|}{Citation Graph} & \multicolumn{8}{c|}{E-Commerce \& Knowledge Graph} & \multicolumn{8}{c|}{Schools} & \multicolumn{2}{c}{} \\ \hline
Method & \multicolumn{2}{c}{Cora} & \multicolumn{2}{c}{Citeseer} & \multicolumn{2}{c|}{Pubmed} & \multicolumn{2}{c}{History} & \multicolumn{2}{c}{Children} & \multicolumn{2}{c}{Sportsfit} & \multicolumn{2}{c|}{Wikics} & \multicolumn{2}{c}{Cornell} & \multicolumn{2}{c}{Texas} & \multicolumn{2}{c}{Wisconsin} & \multicolumn{2}{c|}{Washington} &  &  \\
 & Acc & F1 & Acc & F1 & Acc & \multicolumn{1}{c|}{F1} & Acc & F1 & Acc & F1 & Acc & F1 & Acc & F1 & Acc & F1 & Acc & F1 & Acc & F1 & Acc & F1 & Acc & F1 \\ \hline
Sbert~\cite{reimers2019sentence} & 69.75 & 67.21 & 66.69 & 63.31 & 70.57 & \multicolumn{1}{c|}{71.38} & 53.53 & 20.45 & 22.59 & 20.13 & 43.79 & 38.26 & 59.06 & 56.19 & 63.66 & 54.39 & 64.58 & 49.79 & 62.10 & 52.07 & 63.52 & 48.00 & 7.27 & 7.09 \\
Roberta~\cite{liu2019roberta} & 70.71 & 68.47 & 66.95 & 63.57 & 69.54 & \multicolumn{1}{c|}{70.31} & 55.39 & 21.84 & 24.25 & 22.41 & 41.51 & 36.09 & 59.08 & 56.49 & 61.68 & 51.84 & 62.25 & 49.26 & 60.33 & 49.08 & 60.60 & 45.34 & 7.18 & 7.18 \\
Text-Embedding-3-Large~\cite{openai2024textembedding} & 71.90 & 69.87 & 66.24 & 63.30 & 75.96 & \multicolumn{1}{c|}{75.75} & 50.15 & 19.21 & 24.68 & 24.10 & 58.39 & 53.03 & 61.78 & 58.82 & 81.50 & 70.11 & 75.42 & 63.17 & 73.14 & 63.02 & 66.35 & \textbf{57.69} & 5.36 & 4.45 \\
LLM2Vec~\cite{behnamghader2024llm2vec} & 67.34 & 65.92 & 67.13 & 64.37 & 74.57 & \multicolumn{1}{c|}{74.65} & 53.14 & 19.06 & 25.56 & 24.31 & 57.00 & 52.29 & 62.34 & 58.32 & 81.26 & 69.08 & 76.68 & 63.12 & 73.36 & 62.50 & 65.92 & 53.34 & 5.64 & 5.36 \\ \hline
SBert + NA~\cite{yang2024gl} & 72.49 & 69.90 & 68.66 & 64.75 & 71.26 & \multicolumn{1}{c|}{71.87} & 57.86 & 21.98 & 25.28 & 22.74 & 46.84 & 40.85 & 66.26 & 63.57 & 54.21 & 44.66 & 56.04 & 41.09 & 54.23 & 46.11 & 58.88 & 43.05 & 5.82 & 6.00 \\ \hline
GPT-3.5-turbo~\cite{achiam2023gpt} & 70.11 & 52.11 & 66.83 & 47.58 & 89.75 & \multicolumn{1}{c|}{66.16} & 55.07 & 30.36 & 29.73 & 26.13 & \textbf{67.21} & 54.45 & 65.53 & 51.19 & 45.54 & 39.30 & 56.14 & 32.53 & 58.86 & 46.84 & 51.09 & 35.68 & 5.64 & 8.18 \\
GPT-4o~\cite{hurst2024gpt} & 70.29 & 62.95 & 64.77 & 47.78 & \textbf{89.85} & \multicolumn{1}{c|}{67.39} & 53.30 & \textbf{31.68} & \textbf{30.76} & \textbf{29.20} & 66.35 & 56.22 & 66.10 & 56.04 & 45.54 & 41.92 & 63.10 & 50.51 & 56.60 & 52.54 & 48.90 & 42.54 & 5.91 & 6.36 \\ \hline
UniGLM~\cite{fang2024uniglm} & 45.57 & 43.25 & 52.26 & 48.41 & 70.33 & \multicolumn{1}{c|}{69.78} & 44.24 & 24.84 & 21.48 & 19.17 & 33.46 & 32.99 & 55.05 & 52.08 & 23.03 & 22.06 & 21.39 & 18.90 & 27.16 & 26.45 & 24.01 & 23.08 & 11.36 & 9.91 \\
ZeroG~\cite{li2024zerog} & 60.4 & 56.02 & 50.35 & 45.15 & 74.68 & \multicolumn{1}{c|}{71.75} & 36.55 & 16.84 & 12.72 & 12.61 & 14.27 & 5.33 & 46.74 & 40.86 & 10.47 & 6.46 & 53.48 & 15.95 & 12.66 & 5.02 & 8.3 & 3.07 & 12.27 & 12.73 \\ \hline
DGI~\cite{velivckovic2018deep} & 16.79 & 12.77 & 15.24 & 15.04 & 25.10 & \multicolumn{1}{c|}{19.18} & 20.98 & 3.89 & 2.22 & 1.04 & 7.48 & 3.47 & 14.98 & 4.24 & 14.66 & 10.02 & 11.23 & 9.42 & 12.08 & 6.95 & 20.96 & 14.15 & 13.91 & 14.73 \\
GraphMAE~\cite{hou2022graphmae} & 15.13 & 7.10 & 8.11 & 7.67 & 36.56 & \multicolumn{1}{c|}{34.29} & 36.36 & 5.75 & 7.24 & 1.97 & 30.50 & 6.99 & 8.91 & 4.03 & 23.04 & 14.95 & 17.65 & 11.67 & 23.02 & 11.87 & 24.89 & 13.34 & 15.18 & 15.45 \\ \hline
OFA~\cite{liu2023one} & 20.36 & 16.57 & 41.31 & 33.37 & 28.18 & \multicolumn{1}{c|}{26.62} & 8.25 & 3.48 & 3.05 & 2.29 & 15.18 & 4.7 & 30.77 & 25.22 & 29.84 & 12.62 & 11.77 & 5.87 & 4.8 & 3.44 & 6.04 & 4.28 & 13.91 & 14.73 \\
GOFA~\cite{kong2024gofa} & 71.06 & 70.21 & 65.72 & 64.18 & 74.76 & \multicolumn{1}{c|}{73.00} & 56.25 & 31.57 & 12.15 & 7.73 & 37.87 & 33.19 & \textbf{68.62} & 62.93 & 39.50 & 35.47 & 38.37 & 29.54 & 32.51 & 25.12 & 31.02 & 21.24 & 8.00 & 7.45 \\ \hline
GraphGPT~\cite{tang2024graphgpt} & 17.48 & 12.68 & 13.93 & 12.78 & 42.94 & \multicolumn{1}{c|}{25.68} & 12.31 & 9.15 & 9.94 & 4.24 & 4.53 & 2.44 & 33.59 & 30.21 & 10.18 & 14.71 & 18.48 & 9.85 & 12.35 & 6.32 & 20.64 & 15.79 & 14.55 & 14.64 \\
LLAGA~\cite{chen2024llaga} & 11.62 & 14.42 & 19.52 & 23.34 & 7.56 & \multicolumn{1}{c|}{13.42} & 7.95 & 8.89 & 10.09 & 5.02 & 1.84 & 2.66 & 10.98 & 16.73 & 12.57 & 20.1 & 15.51 & 22.97 & 15.09 & 20.85 & 10.48 & 18.98 & 15.36 & 13.64 \\ \hline
LLM-BP & \textbf{72.59} & \textbf{71.10} & \textbf{69.51} & \textbf{66.29} & 75.55 & \multicolumn{1}{c|}{75.32} & \textbf{59.86} & 22.66 & 24.81 & 22.66 & 61.92 & \textbf{57.51} & 67.75 & 63.53 & 83.28 & 71.80 & \textbf{81.66} & \textbf{65.41} & \textbf{77.75} & \textbf{63.70} & \textbf{73.14} & 57.33 & \textbf{2.27} & 2.55 \\
LLM-BP (appr.) & 71.41 & 70.11 & 68.66 & 65.62 & 76.81 & \multicolumn{1}{c|}{\textbf{76.81}} & 59.49 & 23.02 & 29.40 & 28.45 & 61.51 & 57.09 & 67.96 & \textbf{64.27} & \textbf{84.92} & \textbf{74.19} & 79.39 & 64.63 & 75.65 & 62.53 & 70.04 & 55.53 & 2.45 & \textbf{2.27} \\ \hline
\end{tabular}}
\vspace{-0.2cm}
\caption{Zero-Shot End-to-End Evaluation. `NA' refers to neighborhood embedding aggregation.}
\label{tab:zero_shot}
\end{table*}
\begin{figure*}[t]
    \centering
    \begin{minipage}{0.22\textwidth}
    \captionsetup{labelformat=empty}
        \centering
        \includegraphics[width=\textwidth]{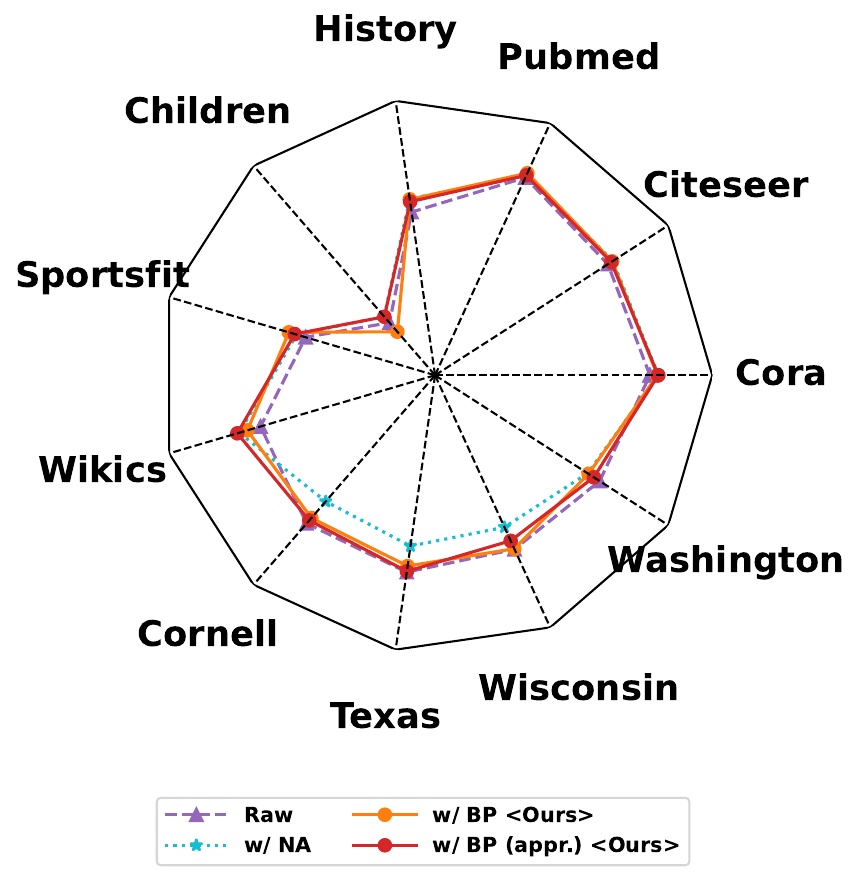}
        \vspace{-0.8cm}
        \caption*{\scriptsize SBert.}
    \end{minipage}
    \begin{minipage}{0.22\textwidth}
    \captionsetup{labelformat=empty}
        \centering
        \includegraphics[width=\textwidth]{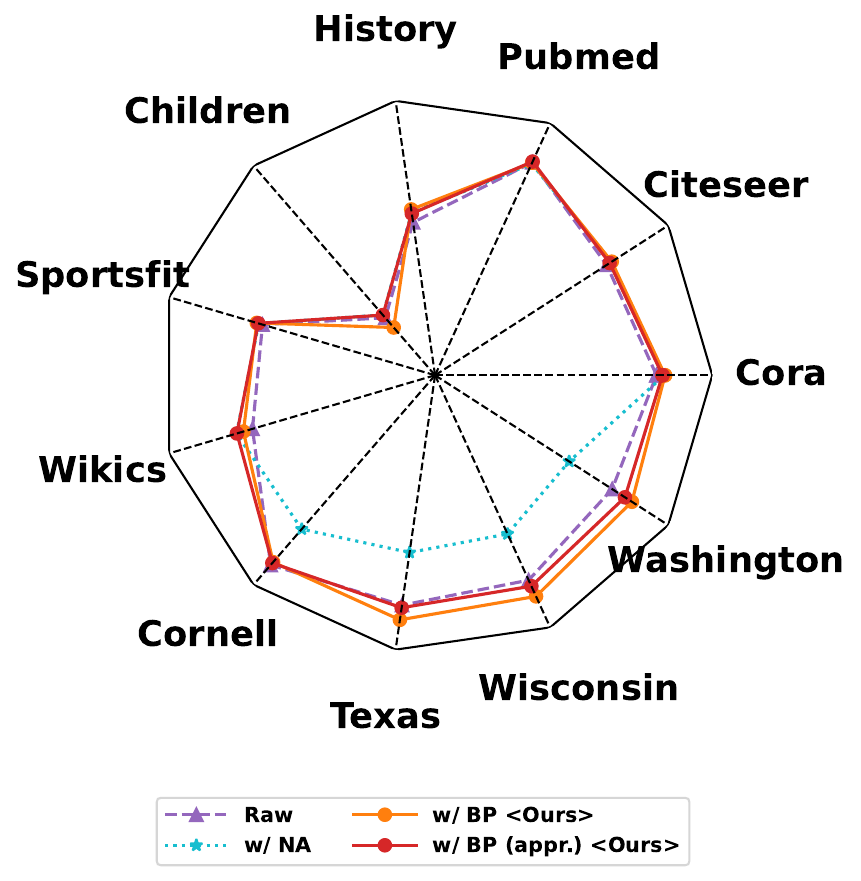}
        \vspace{-0.8cm}
        \caption*{\scriptsize Text-Embedding-3-Large.}
    \end{minipage}
    \begin{minipage}{0.22\textwidth}
    \captionsetup{labelformat=empty}
        \centering
        \includegraphics[width=\textwidth]{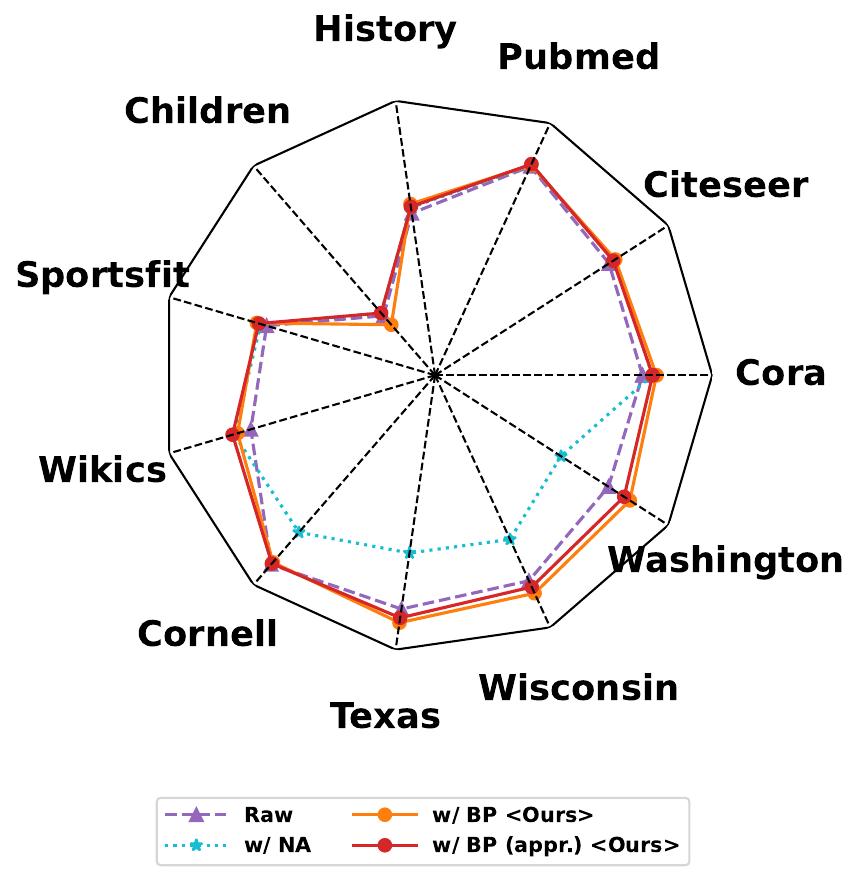}
        \vspace{-0.8cm}
        \caption*{\scriptsize LLM2Vec.}
    \end{minipage}
    \begin{minipage}{0.22\textwidth}
    \captionsetup{labelformat=empty}
        \centering
        \includegraphics[width=\textwidth]{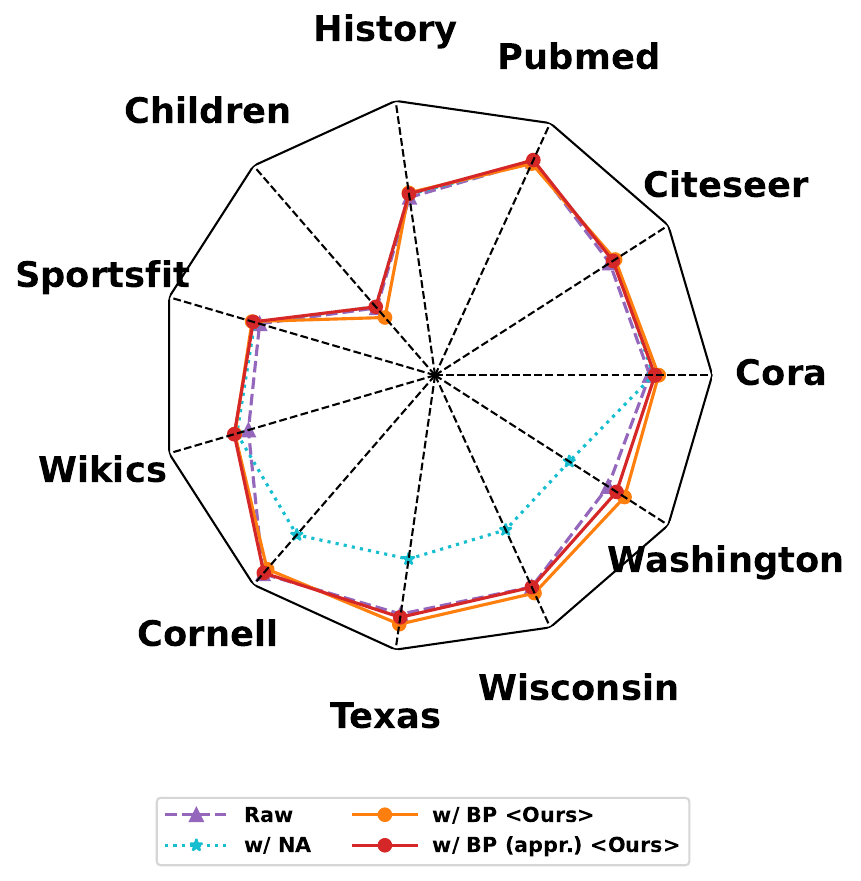}
        \vspace{-0.8cm}
        \caption*{\scriptsize Class-Conditional Encoding (Ours).}
    \end{minipage}
    \vspace{-0.2cm}
    \caption{Experiments on graph information aggregation. `Raw' refers no graph structure usage, `w/ NA' refers to the neighborhood embedding aggregation (NA) proposed in~\cite{yang2024gl}, `w/ BP' refers to the belief propagation following Eq.~\ref{eq:bp}, `w/ BP (appr.)' refers to its simplified linear form that follows Eq.~\ref{eq:bp_appr}.}
    \label{fig:principle_2}
\end{figure*}
$\bullet$ \textbf{Exp.3: LLM Agents for Homophily Level $r$ Estimation} As shown in Fig.\ref{fig:predict_h} (Left), we randomly sample $k$ edges ($k=100$ for large graphs and $k=50$ for small ones), incorporating them into prompts (Fig.\ref{fig:pipe}) for LLM-based estimation of the homophily level $r$ (Sec.\ref{sec:principle2}). We evaluate four LLMs: GPT-4o, GPT-4o-mini\cite{hurst2024gpt}, GPT-3.5-Turbo~\cite{achiam2023gpt}, and Mistral-7B-Instruct v0.3~\cite{jiang2023mistral}. Each model responds to each node pair over five trials, with the final estimate determined by majority voting. Full results are provided in Fig.\ref{fig:predict_h_more} in appendix, demonstrating that GPT-4o-mini and GPT-4o effectively estimate $r$, GPT-3.5-Turbo performs reasonably well, while Mistral-7B-Instruct-v0.3 fails. Balancing accuracy and cost efficiency, we select GPT-4o-mini's estimation (Fig.~\ref{fig:predict_h} Right) for subsequent studies. 

$\bullet$ \textbf{Exp.4: Effectiveness of the BP Algorithm}
Experimental results are presented in Fig.~\ref{fig:principle_2}, where we evaluate the four approaches over various graph structures. Specifically, We compare the BP algorithm (Eq.~\ref{eq:bp}) and its linear approximation (Eq.~\ref{eq:bp_appr}) with vanilla encoders that do not utilize structure (Raw) and the NA baseline. For all the four encoders across all the datasets, the proposed BP algorithm slightly outperforms its linear approximation, and they consistently outperform Raw. Moreover, in most datasets, they also surpass the NA baseline, particularly on heteophilic graphs, where direct neighborhood embedding aggregation negatively affects performance. These results highlight the generalizability of our data modeling approach and the effectiveness of the key-parameter estimation design in BP.

\subsection{End-to-End Evaluation}

$\bullet$ \textbf{Exp.5: Main Results in the Zero-Shot Setting}
The main experimental results are presented in Table~\ref{tab:zero_shot}. Among the baselines, vanilla encoders and LLMs demonstrate strong zero-shot generalization. GPT-3.5-Turbo~\cite{achiam2023gpt} ranks first on the Sportsfit dataset, while GPT-4o~\cite{hurst2024gpt} achieves the best performance on Pubmed and Children.

UniGLM~\cite{fang2024uniglm} and ZeroG~\cite{li2024zerog} perform well in domains aligned with their pre-training, such as citation networks (e.g., ZeroG enhances SBert's performance on Cora, Pubmed, and Wikics). However, both struggle on TAGs with unseen text distributions (e.g., Sportsfit) or novel graph structures (e.g., webpage networks), suggesting that fine-tuned LM encoders may suffer performance degradation on out-of-domain TAGs.
Similarly, graph-SSL methods (DGI~\cite{velivckovic2018deep}, GraphMAE~\cite{hou2022graphmae}) show limited generalization across structural shifts. 

Among multi-task graph foundation models, GOFA achieves strong performance, likely benefiting from a larger pre-training corpus for graph-text alignment~\cite{hu2021ogb, ding2023enhancing} compared to GraphGPT~\cite{tang2024graphgpt} and LLaGA~\cite{chen2024llaga}, which are trained solely on ogbn-arxiv. However, GOFA still requires broader pre-training and instruction fine-tuning to improve generalization under text domain shifts, and its reliance on GNNs may limit effectiveness on heterophilic data.

Notably, LLM-BP and LLM-BP (appr.) achieve the highest average ranking across all datasets on both homophilic and heterophilic graphs. For fine-grained average ranking that distinguish between homophilic and heterophilic graphs, refer to Appendix.~\ref{sec:app_fine_grained_ranking}. Another interesting observation is that when we randomly sample $20c$ nodes to obtain the class embeddings with the help of LLMs following Algorithm.~\ref{alg:llm_bp}, the zero-shot performance of the encoders in this setting is comparable to their performance between $5$-$10$-shot setting as shown in Table.~\ref{tab:few_shot} in the Appendix.
Further comparisons with LLM-GNN~\cite{chen2023label} and TEA-GLM~\cite{wang2024llms} are provided in Appendix~\ref{sec:app_more_baselines}.

$\bullet$ \textbf{Exp.6: Main Results under Few-shot Setting} We conduct the evaluation in $k=1,3,5,10$-shot settings. 
Using 10 different random seeds, we sample the shots from the training set and repeat the experiments 10 times. The experimental results are presented in Table~\ref{tab:few_shot} in Appendix~\ref{sec:app_more_experiment_results}. Across all 
$k$-shot settings, LLM-BP and LLM-BP (appr.) outperform the baseline models.

\section{Discussion and Limitations}

Graph learning tasks often face substantial data constraints compared to other domains, underscoring the importance of establishing fundamental principles that foster model generalization. Our approach exemplifies this by leveraging LLMs to analyze graph data and determine suitable inference strategies, particularly via homophily estimation for belief propagation. While \proj achieves notable success on TAGs for node classification and extends partially to link prediction, it remains a step away from a fully comprehensive graph foundation model that addresses a wider range of graph learning tasks. Nonetheless, the core idea of leveraging LLM-driven graph analysis to guide algorithmic decisions aligned with task-specific inductive biases holds broad potential for future applications.

\section*{Acknowledgements}
H. Wang, S. Liu, R. Wei and P. Li are partially supported by NSF awards IIS-2239565, CCF-2402816, IIS-2428777, PHY-2117997; DOE award DE-FOA-0002785; JPMC faculty awards; Openai Research Credits; and Meta research award.

We extend our sincere gratitude to Hongbin Pei, Zhen Wang, and Jingcheng Cen for their valuable assistance in identifying the raw node text of the heterophilic graphs used in this study.


\section*{Impact Statement} This paper presents work whose goal is to advance the field of 
Machine Learning. There are many potential societal consequences 
of our work, none which we feel must be specifically highlighted here.

\bibliography{example_paper}
\bibliographystyle{icml2025}

\newpage
\appendix
\onecolumn
\section{More Related Works}
\label{sec:app_more_related_works}
\textbf{LLMs for Data Augmentation} annotate pseudo-labels via their advanced zero-shot text classification performance. \textit{E.g.}, LLM-GNN~\cite{chen2023label}, Cella~\cite{zhang2024cost} and ~\cite{hu2024low} propose heuristics to actively select and annotate pseudo-labels for supervised GNN training. ~\cite{pan2024distilling} performs knowledge distillation with LLMs as teachers. ~\cite{yu2023empower,li2024enhancing} generate synthetic node text with LLMs. The performance of these methods depend on the capability of LLM, and may still require relatively high annotating and training cost.

\textbf{LLMs for Graph Property Reasoning} focus on reason graph structure properties (e.g., shortest path, node degree, etc)~\cite{tang2024grapharena, dai2024large, yuan2024gracore, ouyang2024gundam}. Representative works include~\cite{perozzi2024let, chen2024graphwiz, zhang2024can, cao2024graphinsight, wei2024gita}.

\textbf{Tuning LMs/GNNs towards Better Task-Specific Performance} aims to push the limits of task-specific performance on TAGs other than generalization. These methods develop novel techniques to optimize LMs or GNNs for pushing the limits of in-domain performance~\cite{chien2021node, duan2023simteg, he2023harnessing, zhao2022learning, zhu2021textgnn, li2021adsgnn, yang2021graphformers, bi2021leveraging, pang2022improving, zolnai2024stage, yang2021graphformers}.

\textbf{Text embeddings} Generating unified text embeddings is a critical research area with broad applications, including web search, accounting documents~\cite{li2024enhancing2} and question answering. Numerous text encoders~\cite{reimers2019sentence, liu2019roberta, song2020mpnet} based on pre-trained language models have served as the foundation for various embedding models. Recently, decoder-only LLMs have been widely adopted for text embedding tasks~\cite{li2023towards, moreira2024nv} achieving remarkable performance on the Massive Text Embedding Benchmark (MTEB)~\cite{muennighoff2022mteb}. This progress stems from LLM2Vec~\cite{behnamghader2024llm2vec}, which introduces a novel unsupervised approach to transforming decoder-only LLMs into embedding models, including modifications to enable bidirectional attention. Recent findings~\cite{li2024making} suggest that retaining the unidirectional attention mechanism enhances LLM2Vec’s empirical performance.

\section{Experiment Details}

\subsection{Dataset Details}
\label{sec:app_datasets}
\textbf{Meta-Data}
In Table.~\ref{tab:dataset_meta_data}, we show the meta-data of all the eleven datasets used in our experiments.

\begin{table}[h]
\centering
\begin{tabular}{@{}ccccc@{}}
\toprule
 & \begin{tabular}[c]{@{}c@{}}Number\\ of Nodes\end{tabular} & \begin{tabular}[c]{@{}c@{}}Number \\ of  Edges\end{tabular} & \begin{tabular}[c]{@{}c@{}}Number \\ of  Classes\end{tabular} & \begin{tabular}[c]{@{}c@{}}Ground Truth\\ Homophily Ratio\end{tabular} \\ \midrule
Cora & 2708 & 10556 & 7 & 0.809 \\
Citeseer & 3186 & 8450 & 6 & 0.764 \\
Pubmed & 19717 & 88648 & 3 & 0.792 \\
History & 41551 & 503180 & 12 & 0.662 \\
Children & 76875 & 2325044 & 24 & 0.464 \\
Sportsfit & 173055 & 3020134 & 13 & 0.9 \\
Wikics & 11701 & 431726 & 10 & 0.678 \\
Cornell & 191 & 292 & 5 & 0.115 \\
Texas & 187 & 310 & 5 & 0.067 \\
Wisconsin & 265 & 510 & 5 & 0.152 \\
Washington & 229 & 394 & 5 & 0.149 \\ \bottomrule
\end{tabular}
\vspace{-0.cm}
\caption{Meta data of the datasets in this study.}
\label{tab:dataset_meta_data}
\end{table}

\textbf{Dataset Split} For the datasets (all the homophily graphs) that have been used for study in TSGFM~\cite{chen2024text}, we follow their implementation to perform data pre-processing, obtain raw texts and do data split, the introduction to data source can be found at Appendix.D.2 in their original paper, the code can be found at the link \footnote{\url{https://github.com/CurryTang/TSGFM/tree/master?tab=readme-ov-file}}.

As to the heterophily graphs, the four datasets are originally from~\cite{craven1998learning}. We obtain the raw texts from~\cite{yan2023comprehensive}, which can be found from\footnote{\url{https://github.com/sktsherlock/TAG-Benchmark/tree/master}}. As to data split, for zero-shot inference, all the nodes are marked as test data; for few-shot setting, $k$ labeled nodes are randomly sampled per class and the rests are marked as test data.
To the best of our knowledge, the four heterophily graph datasets used in this study are the only graphs that provide raw texts feature.

\subsection{LLM-BP Implementation Details}
\label{sec:app_llm_bp_implementation_details}
\textbf{Infrastructure and Seeds}
All the local experiments run on a server with AMD EPYC 7763 64-Core Processor and eight NVIDIA RTX 6000 Ada GPU cards, methods are mainly implemented with PyTorch~\cite{paszke2019pytorch}, Torch-Geometric~\cite{fey2019fast} and Huggingface Transformers~\cite{wolf2019huggingface}.
To obtain the embeddings, all the encoders that run locally on the server without API calling in this study run with the random seed $42$.

\textbf{Class Embedding}

$\bullet$ \textbf{Zero-Shot Setting:} We uniformly randomly sample $20 c$ nodes per graph, where $c$ denotes the number of classes, we employ GPT-4o~\cite{hurst2024gpt} to infer their labels. With the predictions from LLMs, the sampled nodes form distinct clusters. For each cluster, we take the top-$k$ (10 in the experiments) nodes whose embedding are closest with the cluster center and calculate their average embedding as the class embedding.

We notice that some works directly feed text descriptions into encoders as class embeddings~\cite{yang2024gl, chen2024text}, we find that different encoders can be highly sensitive to variations in text description. Therefore, we adopt the above method to ensure fairness among different encoders.

$\bullet$\textbf{Few-Shot Setting:} We directly take the class embedding as the averaged embeddings of labeled nodes per class.

\textbf{The Task-Adaptive Encoder:} We directly adopt the pre-trained LLM2Vec encoder released by~\cite{li2024making}, which is based on Mistral7B-v0.1~\cite{jiang2023mistral}. We check the pre-training data used in the original paper for aligning LLM decoders with the embedding space, the datasets are mainly for text-retrieval and therefore do not overlap with the TAG datasets adopted in our study. For detailed introduction of the datasets for LLM2Vec pre-training, see Section 4.1 training data in the original paper.
\label{sec:app_task_adaptive_implementation}
The task-adaptive prompting follows the format as:

\textit{
\text{$\langle$ Instruct $\rangle$}\\
``Given the \text{\{task description\}}, classify it into one of the following $k$ classes: \\
\text{\{class labels\}}\\
\text{$\langle$ query$\rangle$}\\
\text{\{raw node texts\}}.''\\
\text{$\langle$ response $\rangle$}\\}

, where the \{task descriptions\} prompts for each dataset is the same as that used for vanilla LLMs, see Table.~\ref{tab:vanilla_llm_task_description} for details.

\textbf{Hyper-Parameters for BP algorithm}
For LLM-BP, we adopt $5$ message-passing layers, for its linear approximation form, we use a single layer.
The temperature hyper-parameter $\tau$ in computing node potential initialization in Eq.~\eqref{eq:node_potential} is set as $0.025$ for LLM-BP and $0.01$ for LLM-BP (appr.) across all the datasets. Attached in Table.~\ref{tab:pred_h} is the homophily ratio $r$ we used (predicted by GPT-4o-mini~\cite{hurst2024gpt}.
\begin{table}[h]
\centering
\resizebox{\textwidth}{!}{\begin{tabular}{@{}cccccccccccc@{}}
\toprule
 & Cora & Citeseer & Pubmed & History & Children & Sportsfit & Wikics & Cornell & Texas & Wisconsin & Washington \\ \midrule
\begin{tabular}[c]{@{}c@{}}Ground Truth \\ Homophony Ratio\end{tabular} & 0.81 & 0.76 & 0.79 & 0.66 & 0.46 & 0.90 & 0.67 & 0.11 & 0.06 & 0.15 & 0.19 \\
\begin{tabular}[c]{@{}c@{}}$r$ predicted by \\ GPT-4o-mini\end{tabular} & 0.70 & 0.81 & 0.81 & 0.73 & 0.35 & 0.81 & 0.52 & 0.05 & 0.04 & 0.06 & 0.02 \\ \bottomrule
\end{tabular}}
\vspace{-0.cm}
\caption{$r$ predicted by GPT-4o-mini, that is used in all the experiments in this study.}
\label{tab:pred_h}
\end{table}

\subsection{Baseline Implementation Details}
\label{sec:app_implementation_details_baseline}
$\bullet$ \textbf{Vanilla Encoders} Vanilla encoders like SBert~\cite{reimers2019sentence}, Roberta~\cite{liu2019roberta} and text-embedding-3-large~\cite{openai2024textembedding} directly encode the raw text of the nodes. LLM2Vec uses the prompts:

\begin{align}
\label{eq:vanilla_llm2vec}
\resizebox{0.43 \textwidth}{!}{$\langle\text{Instruct}\rangle \{\text{task\_description}\} \langle\text{query}\rangle {X_i} \langle\text{response}\rangle$}.
\end{align}, where the $\{\text{task\_description}\}$ for each dataset is provided in Appendix.~\ref{sec:app_prompt_llm2vec_task_description}.

$\bullet$ \textbf{Vanilla LLMs} Prompts for GPT-4o and GPT-3.5-turbo adopts the format as follows:

\label{sec:app_vanilla_LLM_implementation}
\textit{
``role'': ``system''\\
``content'': ``You are a chatbot who is an expert in text classification''\\
``role'': ``user''\\
``content'': ``We have \text{\{task description\}} from the following $k$ categories: \text{\{class labels\}}\\
The text is as follows:\\
\text{\{raw node text\}}\\
Please tell which category the text belongs to:''
}

The \{task description\} for the vanilla LLMs for each class is provided in Appendix.~\ref{sec:app_prompt_vanilla_llm}.

$\bullet$ \textbf{Tuning LM/GNNs}
We adopt the pre-trained UniGLM~\cite{fang2024uniglm} released by the official implementation, which adopts Bert as the encoder, for direct inference.
For ZeroG~\cite{li2024zerog}, we re-implement the method and train it on ogbn-arxiv~\cite{hu2020open} for fair comparison with other baselines.

As to GNNs tuning methods, we pre-train GraphMAE~\cite{hou2022graphmae} and DGI~\cite{velivckovic2018deep} on ogbn-arxiv~\cite{hu2020open}, where the input for both models are from SBert~\cite{reimers2019sentence}, and we follow in implementation in TSGFM~\cite{chen2024text} benchmark.

$\bullet$ \textbf{Multi-Task GFMs} OFA~\cite{liu2023one} is trained on ogbn-arxiv~\cite{hu2020open}. As to GOFA, we directly adopt the model after pre-training~\cite{hu2021ogb, ding2023enhancing} and instruct fine-tuning on ogbn-arxiv~\cite{hu2020open} provided by the authors due to the huge pre-training cost, the zero-shot inference scheme also follows their original implementation.

$\bullet$ \textbf{LLMs with Graph Adapters} Both LLaGA~\cite{chen2024llaga} and GraphGPT~\cite{tang2024graphgpt} are trained on ogbn-arxiv~\cite{hu2020open}, we follow the hyper-parameter setting in their original implementation.

\section{Detailed Derivations}
\label{sec:app_detailed_derivation}
\subsection{Derivation for Eq.~\eqref{eq:message_passing}}
In a node classification task, given a node \(i\), our goal is to minimize the mean-square error (MSE) in predicting the node label under the observations \(\XX\):
\begin{align}
    \min \mathrm{MSE}\bigl(\hat{y}_i\bigr) 
    \;=\; 
    \mathbb{E}\Bigl[\bigl(y_i - \hat{y}_i\bigr)^2 \,\Bigm|\;\XX\Bigr].
\end{align}
The optimal solution \(\hat{y}_i\) is then given by:
\begin{align}
    \hat{y}_i 
    \;=\; 
    \sum_{y_i} y_i \, p\bigl(y_i \mid \XX\bigr),
\end{align}
where the posterior marginal \(p\bigl(y_i \mid \XX\bigr)\) is computed as:
\begin{align}
    p\bigl(y_i \mid \XX\bigr) = \sum_{Y \setminus i} \mathbb{P}\bigl(Y \mid \XX\bigr).
\end{align}

\paragraph{Factorized Posterior under an MRF.}
Assuming a Markov Random Field (MRF) over a graph \(\mathcal{G}=(\mathcal{V}, \mathcal{E})\), the posterior distribution factors as:
\begin{align}
    \mathbb{P}_\mathcal{G}(Y \mid \XX) 
    \;\propto\; 
    \prod_{i \in \mathcal{V}} \varphi_{X_i}\bigl(y_i\bigr)
    \prod_{(i,j) \in \mathcal{E}} \psi_{ij}\bigl(y_i, y_j\bigr),
\end{align}
where the node potential is defined as \(\varphi_{X_i}\bigl(y_i\bigr) = \varphi_{y_i}\bigl(X_i\bigr)\phi_i\bigl(y_i\bigr)\).

\paragraph{General Message-Passing Framework.}
To compute the marginal \(p(y_i \mid \XX)\), the loopy belief propagation (LBP) algorithm iteratively updates messages between nodes. The general message update rule from node \(i\) to node \(j\) at iteration \(k\) is:
\begin{align}
    m_{i \to j}^{(k)}\bigl(y_j\bigr) 
    \;=\; 
    \alpha_{i \to j} \sum_{y_i} \Bigg[\varphi_{X_i}\bigl(y_i\bigr) \psi_{ij}\bigl(y_i, y_j\bigr)
    \prod_{\ell \in \mathcal{N}(i) \setminus j} m_{\ell \to i}^{(k-1)}\bigl(y_i\bigr)\Bigg],
    \label{eq:general_message}
\end{align}
where \(\alpha_{i \to j}\) is a normalization constant ensuring the message sums to 1.

\paragraph{Node Belief Updates.}
The node belief \(p_i^{(k)}\bigl(y_i\bigr)\) at iteration \(k\) is obtained by combining the node potential with incoming messages from all neighbors:
\begin{align}
    p_i^{(k)}\bigl(y_i\bigr) 
    \;=\; 
    \varphi_{X_i}\bigl(y_i\bigr) 
    \prod_{\ell \in \mathcal{N}(i)} m_{\ell \to i}^{(k)}\bigl(y_i\bigr).
    \label{eq:node_belief_update}
\end{align}

\paragraph{Reformulating the Messages.}
Substituting Eq.~\eqref{eq:node_belief_update} into Eq.~\eqref{eq:general_message} simplifies the message-passing equation. The message from node \(i\) to node \(j\) at iteration \(k\) can be rewritten as:
\begin{align}
    m_{i \to j}^{(k)}\bigl(y_j\bigr) 
    \;=\; 
    \alpha_{i \to j} \sum_{y_i} \psi_{ij}\bigl(y_i, y_j\bigr) 
    \frac{p_i^{(k)}\bigl(y_i\bigr)}{m_{j \to i}^{(k-1)}\bigl(y_i\bigr)}.
    \label{eq:message_reformulation}
\end{align}
This reformulation prevents double-counting the contribution of node \(j\) to node \(i\) in the previous iteration.

\paragraph{Log-Space Stability.}
To avoid numerical underflow, the log-space version of the message update is commonly used:
\begin{align}
    \log m_{i \to j}^{(k)}\bigl(y_j\bigr) 
    &\;=\; 
    \mathrm{LSE}_{y_i} \Bigl[\log \psi_{ij}\bigl(y_i, y_j\bigr) + \log p_i^{(k)}\bigl(y_i\bigr) - \log m_{j \to i}^{(k-1)}\bigl(y_i\bigr)\Bigr],
\end{align}
where \(\mathrm{LSE}(\cdot) \equiv \log \sum \exp(\cdot)\).

\paragraph{Summary.}
By iteratively applying these message updates and node belief calculations, LBP provides an approximation for the posterior marginal \(p(y_i \mid \XX)\). The final prediction \(\hat{y}_i\) under the MMSE criterion is:
\begin{align}
    \hat{y}_i 
    \;=\; 
    \sum_{y_i} y_i \, p_i^{(k)}\bigl(y_i\bigr).
\end{align}
This completes the derivation of the message-passing update in Eq.~\eqref{eq:message_passing}.

\section{More Experiment Results}
\label{sec:app_more_experiment_results}

\subsection{Significance Test of Effectiveness of Task-Adaptive Encoding}
\begin{table}[h]
\setlength{\tabcolsep}{1.5pt}
\resizebox{\textwidth}{!}{\begin{tabular}{@{}cccccccccccccccccccccccc@{}}
\toprule
 &  & \multicolumn{2}{c}{Cora} & \multicolumn{2}{c}{Citeseer} & \multicolumn{2}{c}{Pubmed} & \multicolumn{2}{c}{History} & \multicolumn{2}{c}{Children} & \multicolumn{2}{c}{Sportsfit} & \multicolumn{2}{c}{Wikics} & \multicolumn{2}{c}{Cornell} & \multicolumn{2}{c}{Texas} & \multicolumn{2}{c}{Wisconsin} & \multicolumn{2}{c}{Washington} \\ \midrule
\multirow{2}{*}{\begin{tabular}[c]{@{}c@{}}Task-Adaptive Encoding\\ vs.\\ Vanilla LLM2Vec\end{tabular}} & \begin{tabular}[c]{@{}c@{}}90\% CI\\ low, high\end{tabular} & 0.7\% & 1.3\% & -0.2\% & 1.3\% & 1.3\% & 2.3\% & 3.2\% & 5.2\% & 3.3\% & 3.7\% & 1.0\% & 2.6\% & 1.0\% & 1.8\% & 4.0\% & 4.9\% & 1.5\% & 2.7\% & 0.1\% & 0.8\% & -0.2\% & 0.5\% \\ \cmidrule(l){2-24} 
 & P value & \multicolumn{2}{c}{\textbf{1e-8}} & \multicolumn{2}{c}{0.38} & \multicolumn{2}{c}{\textbf{1e-10}} & \multicolumn{2}{c}{\textbf{1e-9}} & \multicolumn{2}{c}{\textbf{3e-59}} & \multicolumn{2}{c}{\textbf{1e-4}} & \multicolumn{2}{c}{\textbf{3e-8}} & \multicolumn{2}{c}{\textbf{7e-18}} & \multicolumn{2}{c}{\textbf{5e-8}} & \multicolumn{2}{c}{\textbf{1e-3}} & \multicolumn{2}{c}{0.82} \\ \midrule
\multirow{2}{*}{\begin{tabular}[c]{@{}c@{}}Task-Adaptive Encoding\\ vs.\\ Text-Embedding-3-Large\end{tabular}} & \begin{tabular}[c]{@{}c@{}}90\% CI\\ low, high\end{tabular} & -0.3\% & -0.2\% & 0.5\% & 1.0\% & -0.3\% & 1.0\% & 6.9\% & 9.1\% & 4.1\% & 4.4\% & 0.7\% & 2.8\% & 1.1\% & 2.0\% & 3.1\% & 4.1\% & 3.1\% & 4.7\% & 0.1\% & 1.2\% & -0.04\% & -0.1\% \\ \cmidrule(l){2-24} 
 & P value & \multicolumn{2}{c}{6e-27} & \multicolumn{2}{c}{\textbf{8e-7}} & \multicolumn{2}{c}{0.51} & \multicolumn{2}{c}{\textbf{7e-21}} & \multicolumn{2}{c}{\textbf{5e-59}} & \multicolumn{2}{c}{\textbf{0.03}} & \multicolumn{2}{c}{\textbf{1e-9}} & \multicolumn{2}{c}{\textbf{2e-25}} & \multicolumn{2}{c}{\textbf{1e-9}} & \multicolumn{2}{c}{\textbf{0.07}} & \multicolumn{2}{c}{1e-13} \\ \bottomrule
\end{tabular}}
\caption{Lower and upper bound of improvement ($\uparrow$) in accuracy of task-adaptive encoding over baselines with 90\% confidence interval, with significance level $p$ ($\downarrow$).}
\label{tab:confidence}
\end{table}

\label{sec:app_confidence}

We conduct significance test on the improvment of task-adaptive encoding over vanilla LLm2Vec~\cite{li2024making} and Text-Embedding-3-Large~\cite{openai2024textembedding} under the zero-shot setting, with results shown in Table.~\ref{tab:confidence}. We replicate experiment for $100$ times with random seeds from $42$ to $141$ and obtain classification accuracy of each method. To check normality, we first apply Shapiro-Wilk test~\cite{SHAPIRO1965}. If the data follows a normal distribution, we perform a Paired-t test~\cite{student1908probable}; otherwise, we use Wilcoxon Signed-Rank test~\cite{wilcoxon1992individual}, with packages from SciPy~\cite{2020SciPy-NMeth}.
The lower and upper bounds under $90\%$ confidence interval are estimated with bootstrap algorithm~\cite{tibshirani1993introduction} to sample $10,000$ times.
Task-adaptive encoding show statistically significant improvement over vanilla LLM2Vec in $9$ out of $11$ dataset and outperforms Text-Embedding-3-Large in $8$ out of $11$ datasets (bolded in the table).

\subsection{Fine-Grained Ranking Results}
\label{sec:app_fine_grained_ranking}
We report the fine-grained ranking results of each method on homophilic and heterophilic graphs. The ranking is shown in Table.~\ref{tab:fine_grain_ranking}.
Across the three sub-categories of graphs, \proj and its approximation algorithm both achieves the top performance.

\begin{table}[h]
\centering
\resizebox{0.5\textwidth}{!}{\begin{tabular}{ccccc}
\hline
Ranking & \multicolumn{2}{c}{Homophilic} & \multicolumn{2}{c}{Heterophilic} \\ \hline
 & Acc & F1 & Acc & F1 \\ \hline
Sbert & 8.57 & 8.00 & 5.00 & 5.50 \\
Roberta & 7.71 & 7.57 & 6.25 & 6.50 \\
Text-Embedding-3-Large & 6.43 & 5.71 & 3.50 & 2.25 \\
LLM2Vec & 6.86 & 6.14 & 3.50 & 4.00 \\
SBert + NA & 4.57 & 5.00 & 8.00 & 7.75 \\
GPT-3.5-turbo & 4.43 & 7.86 & 7.75 & 8.75 \\
GPT-4o & 4.86 & 6.29 & 7.75 & 6.50 \\
UniGLM & 11.00 & 9.43 & 12.00 & 10.75 \\
ZeroG & 11.29 & 11.00 & 14.00 & 15.75 \\
DGI & 15.29 & 15.71 & 15.00 & 15.00 \\
GraphMAE & 14.86 & 15.29 & 12.25 & 13.75 \\
OFA & 14.29 & 14.29 & 15.25 & 16.25 \\
GOFA & 6.71 & 5.71 & 10.25 & 10.50 \\
GraphGPT & 14.43 & 14.86 & 14.75 & 14.25 \\
LLAGA & 15.86 & 14.71 & 14.50 & 11.75 \\
LLM-BP & \textbf{2.86} & 3.14 & \textbf{1.25} & \textbf{1.50} \\
LLM-BP (appr.) & \textbf{2.86} & \textbf{2.29} & 1.75 & 2.25 \\ \hline
\end{tabular}}
\caption{Average ranking in homophilic and homophilic graphs.}
\label{tab:fine_grain_ranking}
\end{table}

\subsection{LLM Agents' Prediction on Homophily Ratio $r$}
\begin{figure*}[h]
    \centering
    \hfill
    \begin{minipage}{0.23\textwidth}
    \captionsetup{labelformat=empty}
        \centering
        \includegraphics[width=\textwidth]{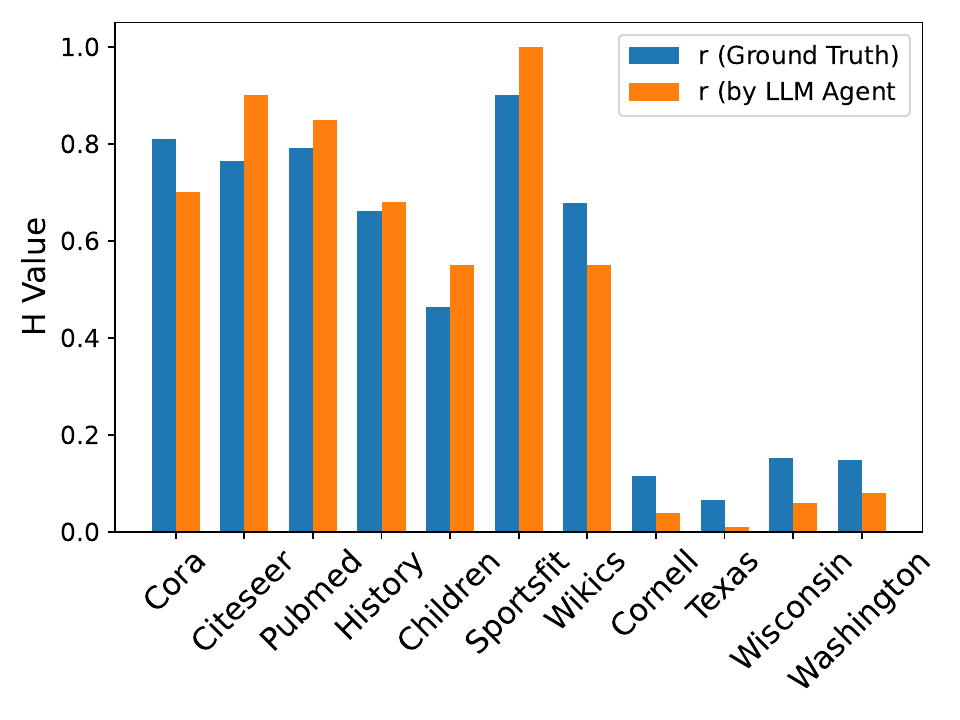}
        \vspace{-0.8cm}
        \caption*{\scriptsize GPT-4o.}
    \end{minipage}
    \hfill
    \begin{minipage}{0.23\textwidth}
    \captionsetup{labelformat=empty}
        \centering
        \includegraphics[width=\textwidth]{camera_ready/figures/pred_h/pred_h_GPT-4o-mini.pdf}
        \vspace{-0.8cm}
        \caption*{\scriptsize GPT-4o-mini.}
    \end{minipage}
    \begin{minipage}{0.23\textwidth}
    \captionsetup{labelformat=empty}
        \centering
        \includegraphics[width=\textwidth]{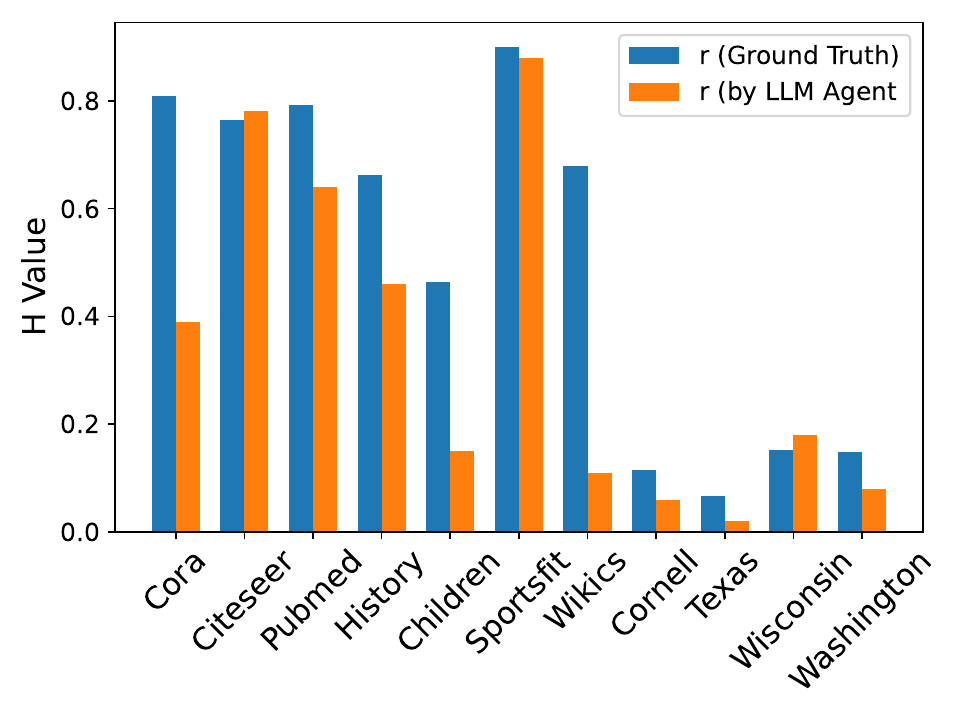}
        \vspace{-0.8cm}
        \caption*{\scriptsize GPT-3.5-turbo.}
    \end{minipage}
    \begin{minipage}{0.23\textwidth}
    \captionsetup{labelformat=empty}
        \centering
        \includegraphics[width=\textwidth]{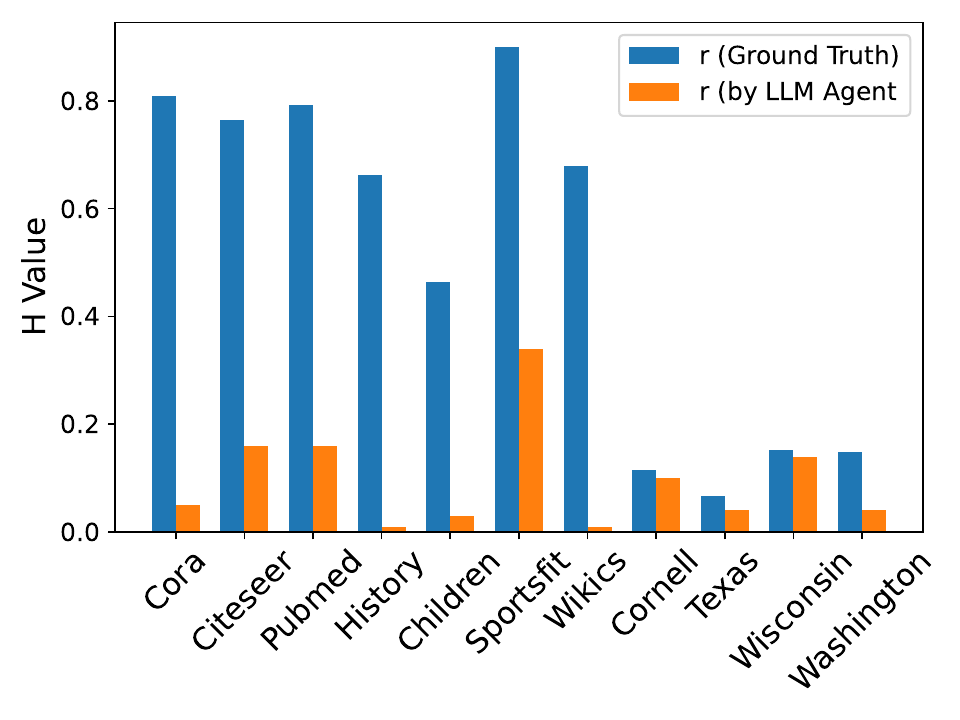}
        \vspace{-0.8cm}
        \caption*{\scriptsize Mistral7B-Instruct-v0.3.}
    \end{minipage}
    \vspace{-0.cm}
    \caption{LLM agents' performance on predicting the homophily constant $r$.}
    \label{fig:predict_h_more}
\end{figure*}
More prediction performance of GPT-4o, GPT-3.5-turbo and Mistral7b-Instruct-v3 are shown in Fig.~\ref{fig:predict_h_more}.


\subsection{Sensitivity Analysis of Sampled Edge Numbers}
\begin{table}[h]
\resizebox{\textwidth}{!}{\begin{tabular}{@{}c|cc|cc|cc|cc|cc|cc|cc@{}}
\toprule
 & \multicolumn{2}{c|}{Cora} & \multicolumn{2}{c|}{Citeseer} & \multicolumn{2}{c|}{Pubmed} & \multicolumn{2}{c|}{Bookhis} & \multicolumn{2}{c|}{Bookchild} & \multicolumn{2}{c|}{Sportsfit} & \multicolumn{2}{c}{Wikics} \\ \midrule
 & value & gap $\downarrow$ & value & gap $\downarrow$ & value & gap $\downarrow$& value & gap $\downarrow$& value & gap $\downarrow$& value & gap $\downarrow$& value & gap $\downarrow$\\ \midrule
ground truth & 0.81 & - & 0.76 & - & 0.79 & - & 0.66 & - & 0.46 & - & 0.90 & - & 0.67 & - \\
100 & 0.70 & 0.11 & 0.81 & 0.05 & 0.81 & 0.02 & 0.73 & 0.07 & 0.35 & 0.11 & 0.81 & 0.09 & 0.52 & 0.15 \\
80 & 0.70 & 0.11 & 0.77 & 0.01 & 0.83 & 0.04 & 0.75 & 0.09 & 0.37 & 0.09 & 0.76 & 0.14 & 0.55 & 0.12 \\
40 & 0.65 & 0.16 & 0.77 & 0.01 & 0.81 & 0.02 & 0.75 & 0.09 & 0.33 & 0.13 & 0.75 & 0.15 & 0.50 & 0.17 \\ \bottomrule
\end{tabular}}
\caption{Sensitivity test of homophily ratio prediction performance with respect to the number of sampled edges. As shown in the left column, $100$, $80$ and $40$ edges are sampled to feed the LLMs to predict homophily ratio $r$.  `value' refers to the predicted $r$, and `gap' is the gap between prediction and ground truth.}
\label{tab:sensitivity_analysis}
\end{table}

We conduct sensitivity analysis on homophily ratio $r$ prediction with respect to the number of sampled edges that feeds to LLM. According to Table.~\ref{tab:sensitivity_analysis}, the homophily ratio prediction performance is stable across the sampled edge numbers from $40$ to $100$.

\subsection{Zero-Shot Comparison with LLM-GNN~\cite{chen2023label} and TEA-GLM~\cite{wang2024llms}}
\label{sec:app_more_baselines}
\begin{table}[h]
\centering
\begin{tabular}{ccccc}
\hline
 & Cora & Citeseer & Pubmed & Wikics \\ \hline
DA-AGE-W & 74.96 & 58.41 & 65.85 & 59.13 \\
DA-RIM-W & 74.73 & 60.80 & 77.94 & 68.22 \\
DA-GraphPart-W & 68.61 & 68.82 & 79.89 & 67.13 \\ \hline
LLM-BP & 72.59 & 69.51 & 75.55 & 67.75 \\
LLM-BP (app.) & 71.41 & 68.66 & 76.81 & 67.96 \\ \hline
\end{tabular}
\caption{Accuracy compared with LLM-GNN, where `DA' denotes the `C-Density' methods proposed in ~\cite{chen2023label}, `-W' refers to the weighted cross-entropy loss function used for training,
AGE~\cite{cai2017active}, RIM~\cite{zhang2021rim}, GraphPart~\cite{ma2022partition} are different graph active learning baselines used in the original paper.}
\label{tab:compare_llm_gnn}
\end{table}
\begin{table}[h]
\centering
\begin{tabular}{ccccc}
\hline
 & Cora & Pubmed & History & Children \\ \hline
TEA-GLM & 20.2 & 84.8 & 52.8 & 27.1 \\ \hline
LLM-BP & 72.59 & 75.55 & 59.86 & 24.81 \\
LLM-BP (app.) & 71.41 & 76.81 & 59.49 & 29.4 \\ \hline
\end{tabular}
\caption{Accuracy compared with TEA-GLM~\cite{wang2024llms}.}
\label{tab:compare_tea-glm}
\end{table}
Here we present the comparison with LLM-GNN~\cite{chen2023label} in Table.~\ref{tab:compare_llm_gnn}. We compare with three different graph active learning heuristics from their original paper. Our training-free methods, LLM-BP and LLM-BP (appr.) achieves top performance on Citeseer and Wikics, while performs comparably with the baselines in Cora and Pubmed. Note that the results of LLM-GNN are from Table. 2 in the original paper.

The comparison with TEA-GLM is shown in Table.~\ref{tab:compare_tea-glm}. Results of TEA-GLM are from Table.1 in their original paper.

\subsection{Experiment Results in Few-Shot Setting}
\label{sec:app_few_shot}
We use $10$ different random seeds from $42$ to $52$ to sample the $k$-shot labeled nodes from training dataset, and report the average accuracy and macro $F1$ score with standard variance. Results are shown in Table.~\ref{tab:few_shot}. Across all the $k$s, our LLM-BP achieves the top ranking performance across all the eleven datasets, exhibiting similar insights with the zero-shot setting.

\begin{table*}[h]
\setlength{\tabcolsep}{2pt}
\resizebox{1.0\textwidth}{!}{\begin{tabular}{cccccccccccccccccccccccccc}
\hline
 &  & \multicolumn{2}{c}{Cora} & \multicolumn{2}{c}{Citeseer} & \multicolumn{2}{c}{Pubmed} & \multicolumn{2}{c}{History} & \multicolumn{2}{c}{Children} & \multicolumn{2}{c}{Sportsfit} & \multicolumn{2}{c}{Wikics} & \multicolumn{2}{c}{Cornell} & \multicolumn{2}{c}{Texas} & \multicolumn{2}{c}{Wisconsin} & \multicolumn{2}{c}{Washington} & \multicolumn{2}{c}{Avg. Rank} \\ \hline
 &  & Acc & F1 & Acc & F1 & Acc & F1 & Acc & F1 & Acc & F1 & Acc & F1 & Acc & F1 & Acc & F1 & Acc & F1 & Acc & F1 & Acc & F1 & Acc & F1 \\ \hline
\multicolumn{26}{c}{\textbf{1-Shot}} \\ \hline
SBert & Raw & 42.8±5.2 & 42.1±5.7 & 42.4±7.7 & 38.8±6.6 & 52.2±6.7 & 51.1±6.6 & \textbf{30.6±7.3} & 14.9±2.5 & 10.5±3.2 & 9.1±1.7 & 17.8±4.4 & 17.2±3.0 & 34.6±3.2 & 31.0±3.1 & 40.4±9.1 & 32.3±6.9 & 26.3±9.4 & 23.5±6.0 & 47.0±7.7 & 35.1±5.0 & 36.0±12.8 & 27.6±8.4 & 9.9 & 10.7 \\
SBert & BP & 43.9±5.1 & 43.0±5.6 & 43.8±8.1 & 39.9±7.0 & 53.6±6.8 & 52.2±6.7 & 33.0±8.2 & 15.5±2.6 & 9.8±2.7 & 8.4±1.4 & 18.7±5.1 & 18.3±3.5 & 35.9±3.5 & 32.0±3.5 & 41.3±8.9 & 33.4±6.5 & 27.9±8.1 & 25.6±5.7 & 47.8±8.4 & 35.8±5.2 & 35.9±11.4 & 27.9±7.3 & 8.5 & 8.6 \\
SBert & BP (appr.) & 43.1±4.9 & 42.4±5.4 & 42.8±7.7 & 39.1±6.6 & 52.5±6.7 & 51.3±6.6 & 31.2±7.6 & 15.1±2.6 & 10.6±3.3 & 9.2±1.6 & 18.0±4.5 & 17.4±3.1 & 35.2±3.3 & 31.4±3.2 & 40.9±9.0 & 32.9±6.9 & 26.7±9.1 & 24.1±5.9 & 47.3±7.9 & 35.2±5.0 & 36.0±12.3 & 27.8±8.1 & 8.9 & 9.5 \\
T-E-3-Large & Raw & 47.2±5.4 & 46.3±5.7 & 40.2±6.3 & 37.0±5.5 & 55.2±7.3 & 52.8±7.2 & 22.9±7.1 & 12.5±1.9 & 13.3±2.7 & 12.0±1.4 & 33.1±6.6 & 30.6±4.3 & 40.3±4.6 & 36.4±3.4 & 55.4±8.4 & 46.5±7.0 & 50.8±7.7 & 42.5±6.4 & 58.6±5.2 & 47.8±4.3 & 44.2±16.9 & 36.1±10.6 & 7.2 & 7.1 \\
T-E-3-Large & BP & \textbf{49.1±5.3} & \textbf{48.0±5.5} & 41.6±7.0 & 38.0±6.1 & 55.8±8.6 & 52.7±8.5 & 24.1±8.0 & 13.1±2.1 & 12.1±2.3 & 10.8±1.3 & 34.7±7.4 & 32.1±4.8 & \textbf{42.4±5.0} & \textbf{38.0±3.7} & 59.6±6.8 & 50.7±6.6 & 52.9±8.2 & 44.5±6.5 & 59.8±5.6 & 49.0±4.5 & 45.5±15.8 & 37.9±10.4 & 5.2 & 5.1 \\
T-E-3-Large & BP (appr.) & 48.4±5.2 & 47.4±5.6 & 40.7±6.6 & 37.4±5.7 & 55.4±7.7 & 52.8±7.5 & 23.3±7.4 & 12.7±2.0 & \textbf{13.4±2.7} & 12.2±1.4 & 33.5±6.8 & 30.9±4.4 & 41.3±4.8 & 37.3±3.5 & 57.0±7.3 & 47.7±6.8 & 51.8±8.0 & 43.4±6.6 & 59.0±5.0 & 48.5±4.5 & 44.3±16.6 & 36.3±10.2 & 6.1 & 6.0 \\
LLM2Vec & Raw & 39.8±6.6 & 38.0±6.9 & 47.9±9.5 & 42.2±7.7 & 54.9±7.7 & 52.5±7.9 & 24.2±12.9 & 12.8±3.6 & 10.0±1.4 & 9.8±1.3 & 27.6±5.6 & 26.1±4.9 & 33.9±4.1 & 31.3±3.8 & 50.1±12.5 & 40.8±8.8 & 50.9±12.0 & 40.9±6.5 & 61.1±7.7 & 48.7±5.6 & 50.0±20.8 & 36.5±10.4 & 8.5 & 8.7 \\
LLM2Vec & BP & 40.5±6.7 & 38.5±7.0 & 48.9±10.7 & 42.7±8.6 & 54.6±9.1 & 51.5±9.6 & 25.7±14.6 & 13.4±4.1 & 9.4±1.3 & 8.8±1.1 & 29.2±6.3 & 27.8±5.5 & 34.3±4.5 & 31.7±4.2 & 53.0±9.9 & 43.8±6.6 & 52.9±11.2 & 42.5±6.7 & 64.2±7.4 & 51.9±4.5 & 49.4±18.3 & 38.0±9.4 & 7.4 & 7.2 \\
LLM2Vec & BP (appr.) & 40.3±6.6 & 38.4±7.0 & 48.6±10.0 & 42.6±8.1 & 55.2±8.0 & 52.8±8.2 & 25.0±13.7 & 13.3±3.8 & 10.1±1.5 & 10.0±1.4 & 28.0±5.8 & 26.7±5.1 & 34.6±4.6 & 31.9±4.1 & 51.7±11.2 & 42.3±7.4 & 52.1±11.3 & 42.0±6.2 & 62.5±7.2 & 49.8±4.7 & 49.7±20.2 & 36.8±10.0 & 7.4 & 7.1 \\
LLM-BP (ours) & Raw & 43.5±5.9 & 42.4±6.1 & 53.0±8.3 & 47.3±6.6 & 57.8±7.1 & \textbf{54.8±8.8} & 28.7±11.4 & 17.2±3.7 & 13.4±3.1 & 14.7±2.3 & 36.4±7.7 & 35.3±7.3 & 38.4±7.8 & 35.1±5.4 & 57.5±12.4 & 47.9±10.3 & 60.0±12.7 & 48.7±6.7 & 69.5±10.6 & 54.5±6.6 & 53.4±19.4 & 39.0±10.1 & 3.8 & 3.5 \\
LLM-BP (ours) & BP & 46.3±6.8 & 44.4±7.2 & \textbf{54.4±8.9} & \textbf{48.4±7.1} & \textbf{58.2±8.1} & 54.2±10.5 & 30.1±12.9 & 17.7±4.0 & 12.5±2.5 & 13.2±2.0 & \textbf{37.7±8.2} & \textbf{36.6±7.7} & 39.3±9.2 & 35.7±6.5 & \textbf{64.8±7.4} & \textbf{54.2±7.6} & \textbf{63.0±11.3} & \textbf{51.3±6.6} & \textbf{73.6±8.8} & \textbf{59.3±6.4} & 53.6±17.6 & 42.3±10.0 & \textbf{2.3} & \textbf{2.0} \\
LLM-BP (ours) & BP (appr.) & 44.7±6.4 & 43.5±6.4 & 53.7±8.5 & 47.9±6.8 & 57.7±7.4 & 54.5±9.2 & 29.5±12.2 & \textbf{17.7±3.9} & 13.5±3.2 & \textbf{14.8±2.4} & 36.9±7.9 & 35.8±7.5 & 39.3±8.5 & 36.0±6.1 & 61.0±10.8 & 50.8±9.7 & 62.1±11.9 & 50.5±6.5 & 71.3±9.9 & 56.6±6.3 & \textbf{54.1±19.1} & \textbf{40.8±10.6} & 2.7 & 2.4 \\ \hline
\multicolumn{26}{c}{\textbf{3-Shot}} \\ \hline
SBert & Raw & 57.6±5.2 & 56.8±5.3 & 57.3±4.0 & 53.2±3.7 & 62.1±5.1 & 62.7±4.7 & 42.8±2.2 & 21.9±2.4 & 13.6±2.2 & 13.1±0.9 & 31.7±3.8 & 29.3±2.8 & 47.2±4.4 & 44.4±3.8 & 51.8±5.0 & 41.4±4.8 & 49.5±7.3 & 36.1±4.5 & 54.9±6.1 & 43.1±4.7 & 49.4±11.2 & 39.1±6.3 & 11.1 & 11.4 \\
SBert & BP & 58.7±5.3 & 57.9±5.3 & 58.6±4.5 & 54.3±4.2 & 63.7±6.1 & 64.3±5.6 & 47.0±2.4 & 23.4±2.4 & 12.5±1.9 & 11.8±0.8 & 35.1±4.4 & 32.6±3.1 & 49.5±4.5 & 46.4±4.0 & 52.2±4.7 & 42.4±4.6 & 48.9±6.9 & 36.6±4.7 & 55.0±5.9 & 43.4±4.8 & 50.2±9.5 & 39.9±5.8 & 9.6 & 9.5 \\
SBert & BP (appr.) & 58.4±5.1 & 57.6±5.2 & 57.7±4.2 & 53.6±3.9 & 62.5±5.1 & 63.1±4.7 & 43.8±2.2 & 22.3±2.5 & 13.7±2.2 & 13.3±1.0 & 32.3±3.9 & 29.9±2.8 & 48.2±4.5 & 45.4±3.9 & 51.6±4.5 & 41.5±4.6 & 49.7±7.3 & 36.4±4.4 & 54.7±5.7 & 42.9±4.2 & 49.8±10.3 & 39.4±5.9 & 10.4 & 10.5 \\
T-E-3-Large & Raw & 64.1±5.6 & 62.8±4.8 & 55.1±4.3 & 51.8±3.9 & 70.6±4.8 & 69.9±4.9 & 38.9±4.7 & 21.0±1.9 & 18.0±2.5 & 17.9±1.6 & 54.0±4.6 & 49.9±2.8 & 54.8±4.3 & 51.4±4.0 & 72.8±3.3 & 63.2±2.2 & 77.9±7.7 & 67.8±11.6 & 69.8±8.1 & 58.9±6.2 & 61.1±13.0 & 49.3±6.5 & 7.5 & 7.9 \\
T-E-3-Large & BP & \textbf{66.3±5.8} & \textbf{64.8±4.9} & 57.5±4.8 & 53.8±4.3 & \textbf{72.4±5.6} & \textbf{71.6±5.7} & 42.4±6.0 & 22.3±2.3 & 16.2±2.1 & 15.9±1.3 & 56.3±5.0 & 52.2±3.1 & 57.0±4.4 & 53.5±4.2 & 73.3±3.7 & 63.6±2.9 & 79.5±7.3 & 69.8±12.2 & 70.9±8.3 & 60.1±6.2 & 63.8±9.5 & 51.4±5.0 & 5.5 & 5.3 \\
T-E-3-Large & BP (appr.) & 65.6±5.8 & 64.1±4.9 & 56.0±4.4 & 52.6±4.0 & 71.1±5.0 & 70.4±5.1 & 40.1±5.2 & 21.5±2.0 & 18.3±2.5 & 18.2±1.6 & 54.7±4.7 & 50.6±2.9 & 56.2±4.4 & 52.8±4.1 & 73.2±3.5 & 63.6±2.9 & 78.7±7.5 & 68.8±11.8 & 70.5±8.4 & 59.4±6.3 & 62.0±12.0 & 50.0±5.8 & 6.3 & 6.5 \\
LLM2Vec & Raw & 56.9±4.9 & 56.5±4.9 & 62.3±4.2 & 58.1±3.7 & 66.6±6.4 & 66.9±5.9 & 45.4±9.2 & 24.2±3.4 & 18.1±3.4 & 17.8±1.1 & 49.5±3.6 & 46.9±2.2 & 51.7±5.1 & 49.5±5.2 & 72.6±5.4 & 61.9±6.9 & 75.0±7.1 & 71.5±8.3 & 70.7±6.1 & 59.5±4.7 & 63.3±10.2 & 49.6±3.9 & 8.2 & 8.0 \\
LLM2Vec & BP & 59.6±5.3 & 59.0±5.3 & 64.6±4.5 & 59.9±3.9 & 67.3±7.5 & 67.4±6.7 & 48.9±10.5 & 25.8±4.0 & 16.4±2.8 & 15.6±1.0 & 52.8±3.9 & 50.1±2.5 & 53.9±6.0 & 51.5±5.7 & 74.1±5.2 & 64.2±6.5 & 79.6±4.6 & 74.5±8.1 & 72.6±5.0 & 60.7±4.0 & 65.0±10.0 & 51.7±4.1 & 5.5 & 5.5 \\
LLM2Vec & BP (appr.) & 59.2±5.0 & 58.7±4.9 & 63.7±4.3 & 59.2±3.7 & 67.1±6.6 & 67.4±6.0 & 47.4±9.8 & 25.4±3.7 & 18.7±3.7 & 18.3±1.1 & 50.9±3.7 & 48.2±2.3 & 54.0±5.5 & 51.8±5.3 & 74.1±5.1 & 64.0±6.4 & 77.6±5.2 & 73.2±7.4 & 72.2±5.6 & 60.6±4.6 & 64.6±10.1 & 50.8±4.1 & 6.1 & 5.9 \\
LLM-BP (ours) & Raw & 60.0±4.0 & 59.2±3.9 & 64.6±4.4 & 60.3±3.6 & 68.7±5.2 & 69.0±4.8 & 59.8±6.0 & 32.2±3.0 & 22.3±3.3 & 22.3±1.5 & 58.2±4.0 & 55.7±2.6 & 55.1±4.9 & 53.5±4.6 & 79.9±3.8 & 72.2±5.7 & 82.5±5.4 & 79.0±6.2 & 83.0±4.0 & 72.3±4.6 & 71.4±12.5 & 58.3±6.0 & 3.6 & 3.5 \\
LLM-BP (ours) & BP & 64.3±4.6 & 63.1±4.3 & \textbf{66.5±4.8} & \textbf{61.8±4.0} & 69.7±6.2 & 69.7±5.8 & \textbf{62.8±6.1} & \textbf{33.4±3.2} & 20.3±2.9 & 19.6±1.4 & \textbf{60.2±4.2} & \textbf{57.8±2.8} & \textbf{57.8±5.6} & \textbf{56.0±5.2} & 81.1±4.0 & \textbf{73.6±5.5} & \textbf{85.4±3.5} & 80.4±7.0 & 84.0±4.3 & \textbf{73.3±4.8} & 71.9±10.6 & \textbf{60.1±5.7} & \textbf{1.9} & \textbf{1.7} \\
LLM-BP (ours) & BP (appr.) & 61.7±4.2 & 60.9±4.0 & 65.7±4.8 & 61.3±3.9 & 69.2±5.5 & 69.5±5.1 & 61.6±5.9 & 33.2±3.0 & \textbf{22.7±3.5} & \textbf{22.7±1.6} & 59.2±4.1 & 56.7±2.7 & 57.4±4.7 & 55.8±4.5 & \textbf{81.3±4.0} & 73.6±5.7 & 84.6±4.7 & \textbf{81.3±6.0} & \textbf{84.3±4.1} & 73.2±5.2 & \textbf{72.1±11.8} & 59.1±6.2 & 2.2 & 2.4 \\ \hline
\multicolumn{26}{c}{\textbf{5-Shot}} \\ \hline
SBert & Raw & 61.4±3.5 & 60.8±3.2 & 61.4±4.6 & 57.4±4.2 & 66.5±4.8 & 67.3±4.3 & 46.7±4.5 & 25.3±2.5 & 16.3±2.3 & 15.7±0.9 & 39.1±3.1 & 36.0±1.8 & 50.5±2.5 & 48.8±2.1 & 55.9±3.6 & 45.4±3.7 & 56.1±6.9 & 40.5±5.7 & 59.8±4.5 & 45.5±4.3 & 56.1±7.6 & 44.1±3.2 & 11.5 & 11.6 \\
SBert & BP & 62.7±4.0 & 62.0±3.7 & 62.6±4.7 & 58.3±4.1 & 67.9±5.7 & 68.7±4.9 & 50.8±5.2 & 26.9±2.6 & 14.7±1.9 & 13.9±0.8 & 43.4±3.6 & 40.2±2.1 & 52.8±2.4 & 50.9±2.1 & 57.2±3.6 & 46.4±3.7 & 55.8±7.0 & 41.6±5.5 & 60.0±4.0 & 45.5±3.5 & 53.8±6.1 & 43.1±2.5 & 10.2 & 10.2 \\
SBert & BP (appr.) & 62.2±3.7 & 61.5±3.4 & 61.8±4.6 & 57.8±4.2 & 67.0±4.8 & 67.8±4.3 & 47.8±4.7 & 25.9±2.5 & 16.6±2.4 & 16.0±0.9 & 40.0±3.2 & 36.9±1.9 & 51.6±2.5 & 49.9±2.0 & 55.9±3.5 & 45.5±3.8 & 56.5±7.0 & 41.2±5.8 & 59.9±4.1 & 45.7±4.0 & 55.1±7.0 & 43.7±2.9 & 10.6 & 10.6 \\
T-E-3-Large & Raw & 69.4±3.1 & 68.4±2.7 & 60.9±4.4 & 57.3±3.6 & 74.8±5.0 & 74.4±4.9 & 49.3±5.0 & 26.9±2.9 & 21.7±3.0 & 22.0±1.1 & 61.7±4.9 & 57.3±2.4 & 59.6±3.1 & 56.9±3.2 & 75.0±4.4 & 67.7±3.3 & 84.2±3.0 & 81.2±2.7 & 74.5±7.7 & 61.9±7.5 & 64.1±10.3 & 54.6±4.2 & 7.5 & 7.4 \\
T-E-3-Large & BP & \textbf{70.6±3.1} & \textbf{69.5±2.7} & 63.3±4.6 & 59.3±3.8 & \textbf{75.6±5.5} & \textbf{75.0±5.6} & 53.4±5.7 & 28.3±3.1 & 19.5±2.3 & 19.4±0.9 & 63.9±5.4 & 59.6±2.6 & 62.0±3.3 & 59.1±3.6 & 75.3±4.1 & 67.9±3.1 & 84.7±2.6 & 80.6±6.2 & 76.3±7.2 & 63.3±6.5 & 65.4±9.1 & 56.2±4.6 & 5.4 & 5.5 \\
T-E-3-Large & BP (appr.) & 70.5±3.4 & 69.4±2.9 & 61.9±4.4 & 58.1±3.7 & 75.4±5.2 & 74.9±5.2 & 50.9±5.3 & 27.6±3.1 & 22.1±3.1 & 22.4±1.2 & 62.5±5.1 & 58.1±2.5 & 61.0±3.2 & 58.3±3.3 & 75.4±4.2 & 68.3±3.1 & 84.2±2.9 & 81.3±2.6 & 75.6±7.6 & 62.7±7.3 & 64.6±10.1 & 54.8±4.7 & 6.1 & 5.8 \\
LLM2Vec & Raw & 61.6±3.6 & 61.1±3.4 & 64.6±5.0 & 61.0±4.5 & 71.4±5.9 & 71.8±5.4 & 54.4±7.0 & 29.8±2.9 & 22.5±2.9 & 22.1±1.0 & 56.8±4.5 & 54.4±2.1 & 58.5±3.6 & 56.4±4.2 & 75.5±5.6 & 68.3±5.1 & 81.8±3.8 & 77.6±2.7 & 78.4±5.0 & 64.3±4.4 & 70.3±8.0 & 55.6±3.6 & 7.5 & 7.7 \\
LLM2Vec & BP & 64.0±3.4 & 63.3±3.4 & 66.9±5.3 & 62.9±4.7 & 72.9±6.8 & 73.2±6.2 & 58.5±7.4 & 31.3±3.0 & 20.3±2.5 & 19.1±0.8 & 60.2±5.5 & 57.6±2.5 & 61.7±3.9 & 59.1±4.4 & 76.2±6.0 & 69.2±6.0 & 83.8±2.5 & 80.5±2.1 & 81.0±4.3 & 67.0±3.9 & 71.1±8.3 & 57.7±4.2 & 5.5 & 5.3 \\
LLM2Vec & BP (appr.) & 64.0±3.5 & 63.3±3.5 & 65.9±5.1 & 62.0±4.6 & 72.0±6.1 & 72.4±5.6 & 56.7±7.1 & 31.1±3.1 & 23.2±3.1 & 22.7±1.0 & 58.3±4.8 & 55.9±2.2 & 61.0±3.9 & 58.9±4.4 & 76.3±6.0 & 69.1±5.8 & 83.0±3.4 & 79.4±2.3 & 80.6±4.8 & 66.6±4.2 & 70.7±8.5 & 56.5±4.0 & 5.9 & 6.0 \\
LLM-BP (ours) & Raw & 65.1±2.8 & 64.4±3.0 & 65.5±5.4 & 61.9±4.7 & 73.6±6.2 & 73.9±5.9 & 65.2±4.8 & 37.2±3.5 & 26.5±2.5 & 26.7±1.1 & 64.3±5.4 & 61.3±2.3 & 62.2±3.6 & 60.4±3.5 & 82.8±3.1 & 76.5±4.8 & 87.6±3.9 & 85.9±2.6 & 87.3±3.4 & 76.1±4.9 & \textbf{77.5±8.8} & 63.5±4.1 & 3.5 & 3.6 \\
LLM-BP (ours) & BP & 69.5±3 & 68.3±3.3 & \textbf{67.4±5.8} & \textbf{63.3±5.1} & 74.0±6.5 & 74.1±6.2 & \textbf{68.4±4.8} & \textbf{38.7±3.7} & 24.3±2.2 & 23.5±0.9 & \textbf{66.6±5.6} & \textbf{63.6±2.4} & \textbf{65.2±3.7} & \textbf{63.1±3.7} & 83.3±3.1 & 77.6±4.8 & \textbf{88.2±2.8} & 86.3±2.6 & \textbf{88.4±2.5} & \textbf{76.8±4.0} & 75.8±7.7 & 64.0±4.7 & \textbf{1.9} & \textbf{1.9} \\
LLM-BP (ours) & BP (appr.) & 67.0±2.7 & 66.0±2.8 & 66.7±5.8 & 62.8±5.0 & 73.9±6.2 & 74.1±5.9 & 66.9±4.7 & 38.4±3.6 & \textbf{26.9±2.7} & \textbf{27.0±1.1} & 65.5±5.5 & 62.5±2.4 & 64.6±3.4 & 62.8±3.5 & \textbf{83.6±2.9} & \textbf{77.6±4.4} & 87.9±3.7 & \textbf{86.4±2.6} & 88.4±2.8 & 76.6±4.9 & 77.4±9.0 & \textbf{64.8±5.4} & 2.4 & 2.4 \\ \hline
\multicolumn{26}{c}{\textbf{10-shot}} \\ \hline
SBert & Raw & 69.8±3.3 & 68.5±3.1 & 67.0±2.0 & 63.0±1.8 & 68.9±4.1 & 69.3±4.6 & 57.3±2.7 & 31.2±0.9 & 20.2±1.8 & 19.7±0.9 & 46.6±3.2 & 43.5±2.0 & 59.5±2.1 & 58.1±1.8 & 61.7±2.9 & 48.1±4.6 & 60.5±3.7 & 47.2±6.6 & 66.1±3.4 & 49.5±4.0 & 62.8±4.1 & 46.4±3.6 & 11.3 & 11.2 \\
SBert & BP & 70.6±3.2 & 69.2±2.9 & 68.3±1.9 & 64.1±1.7 & 70.5±4.9 & 70.8±5.5 & 61.6±2.9 & 33.2±1.2 & 18.0±1.4 & 17.3±0.8 & 51.6±3.9 & 48.3±2.3 & 62.4±2.2 & 60.8±1.9 & 62.0±2.7 & 48.8±4.7 & 60.0±3.3 & 45.5±4.8 & 66.2±4.0 & 49.4±3.8 & 58.5±4.3 & 45.3±3.7 & 9.8 & 10.1 \\
SBert & BP (appr.) & 70.3±3.2 & 68.9±3.1 & 67.5±1.8 & 63.4±1.7 & 69.3±4.3 & 69.7±4.8 & 58.5±2.8 & 31.9±1.0 & 20.6±1.9 & 20.1±1.0 & 47.8±3.4 & 44.6±2.0 & 60.8±2.1 & 59.5±1.8 & 62.0±2.6 & 48.7±4.6 & 60.4±3.6 & 46.6±6.2 & 66.4±3.4 & 49.6±4.4 & 61.2±4.1 & 45.8±3.7 & 10.4 & 10.5 \\
T-E-3-Large & Raw & 74.9±1.2 & 73.7±1.4 & 65.7±1.8 & 61.9±1.5 & 78.3±4.2 & 78.1±4.0 & 59.8±6.5 & 33.0±2.1 & 27.1±2.3 & 26.9±1.4 & 68.1±3.8 & 63.1±2.1 & 68.8±1.9 & 66.6±1.7 & 80.1±2.1 & 71.6±3.4 & 86.5±3.2 & 84.2±2.9 & 81.8±3.5 & 68.3±6.5 & 74.7±4.2 & 60.0±4.6 & 7.5 & 7.6 \\
T-E-3-Large & BP & \textbf{76.6±1.3} & \textbf{75.2±1.5} & 68.3±2.0 & 64.0±1.7 & \textbf{79.1±4.4} & \textbf{78.7±4.3} & 63.9±6.7 & 34.8±2.2 & 24.0±1.7 & 23.4±1.0 & 70.0±4.1 & 65.2±2.1 & 71.3±2.0 & 68.8±1.8 & 80.5±2.7 & 72.3±4.1 & 87.2±2.6 & 85.4±2.8 & 81.9±2.9 & 68.7±5.2 & 73.7±3.9 & 60.7±3.7 & 5.5 & 5.5 \\
T-E-3-Large & BP (appr.) & 76.4±1.3 & 75.0±1.6 & 66.7±1.9 & 62.7±1.6 & 78.7±4.3 & 78.4±4.2 & 61.3±6.7 & 33.9±2.2 & 27.6±2.4 & 27.3±1.4 & 68.8±3.9 & 63.9±2.1 & 70.2±1.9 & 68.0±1.8 & 80.7±1.7 & 72.2±3.1 & 86.7±3.3 & 84.6±3.2 & 82.0±3.5 & 68.7±6.5 & 74.3±4.2 & 60.0±4.2 & 6.1 & 6.2 \\
LLM2Vec & Raw & 68.1±2.7 & 66.9±3.0 & 68.0±2.8 & 64.5±2.6 & 72.9±4.2 & 73.2±3.9 & 65.9±4.2 & 37.5±1.8 & 27.2±2.8 & 27.0±1.3 & 63.4±3.3 & 60.7±1.6 & 68.8±2.4 & 66.7±2.4 & 80.6±3.6 & 72.9±4.4 & 85.9±3.0 & 82.6±4.4 & 84.5±1.7 & 69.0±5.1 & 81.6±4.5 & 61.4±5.9 & 7.8 & 7.5 \\
LLM2Vec & BP & 70.4±3.0 & 69.3±3.2 & 70.6±2.9 & 66.5±2.7 & 74.9±4.1 & 75.0±3.8 & 69.9±4.2 & 39.6±2.4 & 24.6±2.5 & 23.2±1.1 & 66.2±3.6 & 63.6±1.7 & 71.2±2.6 & 68.8±2.4 & 80.9±3.1 & 73.5±4.2 & 87.3±2.0 & 84.9±2.8 & 85.9±2.4 & 70.3±4.3 & 80.5±4.4 & 63.0±5.3 & 5.3 & 5.2 \\
LLM2Vec & BP (appr.) & 70.0±3.0 & 68.8±3.1 & 69.5±2.8 & 65.7±2.6 & 73.5±4.2 & 73.8±4.0 & 68.2±4.2 & 39.5±2.1 & 28.1±3.1 & 27.8±1.3 & 64.9±3.5 & 62.2±1.7 & 71.2±2.5 & 69.1±2.4 & 81.5±2.9 & 73.9±3.7 & 86.6±2.6 & 83.8±3.3 & 86.0±1.8 & 70.6±4.4 & 81.7±4.3 & 62.9±5.3 & 5.8 & 5.7 \\
LLM-BP (ours) & Raw & 70.4±3.2 & 69.1±3.4 & 69.1±2.6 & 65.5±2.2 & 74.3±4.3 & 74.5±4.2 & 72.3±3.1 & 43.5±2.1 & 30.8±2.0 & 30.4±0.9 & 68.5±3.4 & 65.2±1.4 & 71.4±2.3 & 69.4±2.3 & 85.9±3.2 & 80.0±4.2 & 90.7±2.2 & 88.0±3.6 & 90.7±1.7 & 78.7±5.3 & 85.9±2.7 & 68.8±3.7 & 3.6 & 3.6 \\
LLM-BP (ours) & BP & 74.2±3.1 & 72.5±3.2 & \textbf{71.3±2.8} & \textbf{67.3±2.2} & 75.5±4.2 & 75.6±4.0 & \textbf{74.5±2.9} & \textbf{44.7±2.4} & 28.2±1.8 & 26.7±0.8 & \textbf{70.5±3.5} & \textbf{67.4±1.6} & \textbf{73.7±2.4} & 71.4±2.4 & 86.1±3.1 & 80.8±4.1 & \textbf{90.7±1.8} & 88.1±3.1 & 90.6±1.7 & 76.9±5.2 & 83.1±3.5 & 67.9±3.5 & \textbf{2.4} & 2.6 \\
LLM-BP (ours) & BP (appr.) & 71.9±3.5 & 70.6±3.7 & 70.4±2.6 & 66.7±2.1 & 74.8±4.4 & 75.0±4.2 & 73.6±3.0 & 44.6±2.4 & \textbf{31.4±2.1} & \textbf{30.8±1.0} & 69.7±3.5 & 66.5±1.5 & 73.6±2.2 & \textbf{71.6±2.1} & \textbf{86.8±2.9} & \textbf{81.0±3.8} & 90.7±1.9 & \textbf{88.1±3.0} & \textbf{91.7±1.7} & \textbf{78.8±5.2} & \textbf{85.0±3.3} & \textbf{68.3±4.0} & 2.5 & \textbf{2.3} \\ \hline
\end{tabular}}
\caption{Few-Shot Performance. `T-E-3-Large' is short for Text-Embedding-3-Large.}
\label{tab:few_shot}
\end{table*}
\subsection{Zero-Shot Link Prediction Results}
\begin{table}[h]
\centering
\setlength{\tabcolsep}{2pt}
\resizebox{1.0\textwidth}{!}{\begin{tabular}{c|ccc|cccc}
\hline
 & \multicolumn{3}{c|}{Citation Graph} & \multicolumn{4}{c}{E-Commerce \& Knowledge Graph} \\
 & Cora & Citeseer & Pubmed & History & Children & Sportsfit & Wikics \\ \hline
OFA & 0.492 & -- & 0.481 & 0.431 & 0.484 & 0.517 & -- \\
LLaGA & 0.527 & -- & 0.543 & 0.515 & 0.500 & 0.502 & -- \\
GraphGPT & 0.520 & -- & 0.569 & 0.449 & 0.422 & 0.597 & -- \\
TEA-GLM & 0.586 & -- & 0.689 & 0.579 & 0.571 & 0.553 & -- \\ \hline
SBert & 0.979±0.033 & 0.990±0.001 & 0.979±0.003 & 0.985±0.002 & 0.972±0.030 & 0.975±0.003 & 0.972±0.003 \\
Text-Embedding-3-Large & 0.975±0.003 & 0.989±0.002 & 0.979±0.003 & 0.985±0.001 & 0.980±0.002 & 0.987±0.001 & 0.977±0.002 \\
LLM2Vec & 0.966±0.004 & 0.982±0.002 & 0.970±0.003 & 0.971±0.002 & 0.973±0.003 & 0.975±0.002 & 0.978±0.003 \\ \hline
\end{tabular}}
\caption{Performance on zero-shot link prediction tasks (AUC). Results of baselines are from~\cite{wang2024llms}.}
\label{tab:link_prediction}
\end{table}
\label{sec:app_link_prediction}

For each dataset, We randomly sample \& remove $1000$ edges and $1000$ node pairs from the graph as testing data. A straightforward approach is to compare the cosine similarity between node embeddings to determine the presence of a link. Specifically, we aggregate embeddings for $3$ layers on the incomplete graph and compute the cosine similarity between node representations, achieving better zero-shot performance than LLMs-with-Graph-Adapters methods~\cite{wang2024llms, chen2024llaga, tang2024graphgpt}, as shown in Table.~\ref{tab:link_prediction}. Note that the performance in the table refers to LLM-with-Graph-Adapters that have only been trained on other tasks and never on link prediction tasks.

We leave the design of task-adaptive embeddings and generalized graph structural utilization for link prediction as future work, including task-adaptive encoding prompts.

\section{Prompts}
\subsection{Task Description for Vanilla LLM2Vec without Class Information}
\label{sec:app_prompt_llm2vec_task_description}
Table.~\ref{tab:llm2vec_task_description} shows the task description for vanilla LLM2Vec encoder across all the datasets.
\begin{table}[h]
\small
\centering
\begin{tabular}{@{}cc@{}}
\toprule
Dataset & Task Description \\ \midrule
Cora & Encode the text of machine learning papers: \\
Citeseer & Encode the description or opening text of scientific publications: \\
Pubmed & Encode the title and abstract of scientific publications: \\
History & Encode the description or title of the book: \\
Children & Encode the description or title of the child literature: \\
Sportsfit & Encode the title of a good in sports \& fitness: \\
Wikics & Encode the entry and content of wikipedia: \\
Cornell & Encode the webpage text: \\
Texas & Encode the webpage text: \\
Wisconsin & Encode the webpage text: \\
Washington & Encode the webpage text: \\ \bottomrule
\end{tabular}
\caption{\{Task description\} for vanilla LLM2Vec~\cite{li2024making} encoder. See Eq.~\ref{eq:vanilla_llm2vec} for detailed prompting format.}
\label{tab:llm2vec_task_description}
\end{table}

\subsection{Prompts for Vanilla LLMs}
\label{sec:app_prompt_vanilla_llm}
Table.~\ref{tab:vanilla_llm_task_description} shows tha task description for vanilla LLM decoders.
\begin{table}[h]
\small
\centering
\begin{tabular}{@{}cc@{}}
\toprule
Dataset & Task Description \\ \midrule
Cora & opening text of machine learning papers \\
Citeseer & description or opening text of scientific publications \\
Pubmed & title and abstract of scientific publications \\
History & description or title of the book \\
Children & description or title of the child literature \\
Sportsfit & the title of a good in sports \& fitness \\
Wikics & entry and content of wikipedia \\
Cornell & webpage text \\
Texas & webpage text \\
Wisconsin & webpage text \\
Washington & webpage text \\ \bottomrule
\end{tabular}
\caption{\{Task description\} in the prompts for both vanilla LLM decoders (See Section.~\ref{sec:app_vanilla_LLM_implementation}) and task-adaptive encoder (See Section.~\ref{sec:app_task_adaptive_implementation}).}
\label{tab:vanilla_llm_task_description}
\end{table}


\newpage





\end{document}